\documentclass{article} 

\usepackage[final]{COLM_2026_Template/colm2026_conference}

\usepackage{microtype}
\usepackage{graphicx}
\usepackage{subcaption}
\usepackage[dvipsnames]{xcolor}
\usepackage[linesnumbered,ruled,vlined]{algorithm2e}

\SetCommentSty{mycommfont}
\usepackage{booktabs} 
\usepackage{algorithmic}
\usepackage{wrapfig}

\usepackage{hyperref}
\usepackage{mmstyles}
\usepackage{lineno}

\ifcolmfinal
\fi

\newcommand{\blfootnote}[1]{%
  \begingroup
  \renewcommand{\thefootnote}{}%
  \footnote{#1}%
  \addtocounter{footnote}{-1}%
  \endgroup
}

\usepackage{amssymb}
\usepackage{mathtools}
\usepackage{amsthm}

\usepackage[capitalize,noabbrev]{cleveref}

\usepackage[disable,textsize=tiny]{todonotes}

\hypersetup{
    colorlinks=true,
    linkcolor=red,
    citecolor=Cerulean,
    filecolor=magenta,      
    urlcolor=magenta,
}

\usepackage[capitalize,noabbrev]{cleveref}
\crefname{section}{§}{§§}
\Crefname{section}{§}{§§}

\usepackage{calligra}
\DeclareMathAlphabet{\mathcalligra}{T1}{calligra}{m}{n}

\usepackage{pifont}

\theoremstyle{plain}

\theoremstyle{definition}

\theoremstyle{remark}

\renewcommand{\paragraph}[1]{\vspace{1mm}\noindent\textbf{#1}}

\usepackage[most]{tcolorbox}
\tcbset{
  promptbox/.style={
    top=10pt,
    colback=lightgray!20,
    colframe=Black,
    colbacktitle=NavyBlue,
    enhanced,
    center,
    attach boxed title to top center={yshift=-0.1in,xshift=0.0in},
    boxed title style={boxrule=0pt,colframe=white,},
  }
}
\newtcolorbox{promptbox}[2][]{promptbox, title=#2,#1}
\tcbset{
  takeawaybox/.style={
    top=10pt,
    colback=lightgray!20,
    colframe=Black,
    colbacktitle=BurntOrange,
    enhanced,
    center,
    attach boxed title to top center={yshift=-0.1in,xshift=0.0in},
    boxed title style={boxrule=0pt,colframe=white,},
  }
}
\newtcolorbox{takeawaybox}[2][]{takeawaybox, title=#2,#1}
\tcbset{
  observationbox/.style={
    top=10pt,
    colback=lightgray!20,
    colframe=Black,
    colbacktitle=YellowGreen,
    enhanced,
    center,
    attach boxed title to top center={yshift=-0.1in,xshift=0.0in},
    boxed title style={boxrule=0pt,colframe=white,},
  }
}
\newtcolorbox{observationbox}[2][]{observationbox, title=#2,#1}
\newtcolorbox{AIbox}[2][]{aibox,title=#2,#1}
\usepackage{xspace}

\newcommand\methodname{Intern-S1-MO}

\title{\methodname{}: Long-horizon Reasoning Agent for Olympiad-Level Mathematical Problem Solving}

\author{Yuzhe Gu$^{1,2*}$ \quad Songyang Gao$^{1*}$ \quad Zijian Wu$^{1,3*}$ \quad Lingkai Kong$^{1,2*}$ \quad Wenwei Zhang$^{1*\dag}$ \\
\bf Zhongrui Cai$^{1}$ \quad Fan Zheng$^{4}$ \quad Tianyou Ma$^{1}$ \quad Junhao Shen$^{1,2}$ \quad Haiteng Zhao$^{1}$ \\
\bf Duanyang Zhang$^{5}$ \quad Huilun Zhang$^{6}$ \quad Kuikun Liu$^{1}$ \quad Chengqi Lyu$^{1}$ \\
\bf Yanhui Duan$^{1}$ \quad Chiyu Chen$^{1}$ \quad Ningsheng Ma$^{1}$ \quad Jianfei Gao$^{1}$ \quad Han Lyu$^{1}$ \\
\bf Dahua Lin$^{1,3}$ \quad Kai Chen$^{1\dag}$ \\
$^1$Shanghai AI Laboratory \quad $^2$Shanghai Jiao Tong University\quad \\
$^3$MMLab, The Chinese University of Hong Kong \\
$^4$ICMAT, Spanish National Research Council \\
$^5$The High School Affiliated to Renmin University of China \\
$^6$Ren Hui Academy of Beijing \\
\texttt{\{guyuzhe1116\}@sjtu.edu.cn, \quad \{zhangwenwei,chenkai\}@pjlab.org.cn}
}

\begin{document}

\ifcolmsubmission
\linenumbers
\fi

\maketitle
\blfootnote{$^*$ Equal contribution, $\dag$ Corresponding author}


\begin{abstract}
Large Reasoning Models (LRMs) have expanded the mathematical reasoning frontier through Chain-of-Thought (CoT) techniques and Reinforcement Learning with Verifiable Rewards (RLVR).
However, the performance of LRMs is heavily dependent on the extended reasoning context length. For solving ultra-hard problems like those in the International Mathematical Olympiad (IMO), the required reasoning complexity surpasses the space that an LRM can explore in a single round. Previous works attempt to extend the reasoning context of LRMs but remain prompt-based and built upon proprietary models, lacking systematic structures and training pipelines.
Therefore, this paper introduces Intern-S1-MO, a long-horizon math agent that conducts multi-round hierarchical reasoning, composed of an LRM-based multi-agent system including reasoning, summary, and verification. By maintaining a compact memory in the form of lemmas, Intern-S1-MO can more freely explore the lemma-rich reasoning spaces in multiple reasoning stages, thereby breaking through the context constraints for IMO-level math problems.
Furthermore, we propose OREAL-H, an RL framework for training the LRM using the online explored trajectories to simultaneously bootstrap the reasoning ability of the LRM. 
Experiments show that Intern-S1-MO obtains 26 out of 35 points on the non-geometry problems of IMO2025, matching the performance of silver medalists.
It also surpasses the current advanced LRMs on inference benchmarks such as HMMT2025 and AIME2025.
In addition, it officially participates in CMO2025 and achieves a score of 102/126 under the judgment of human experts, reaching the gold medal level.
\end{abstract}


\section{Introduction}
\label{sec: intro}

Reasoning over complex mathematical problems requires the integration of deductive logic, pattern recognition, and creative problem decomposition. In recent years, large reasoning models (LRMs) have made substantial progress in mathematical reasoning, driven primarily by techniques such as Chain-of-Thought (CoT) \citep{Zhang2022AutomaticCO, Wang2023PlanandSolvePI} and Reinforcement Learning from Verifiable Rewards (RLVR) \citep{Shao2024DeepSeekMathPT, Yue2025DoesRL, Zeng2025SimpleRLZooIA}.
Along with the increasing reasoning capabilities of LRMs, a clear trend is that LRMs are being allocated more thinking budgets for more difficult problems to support exploration of larger solution spaces and trial-and-error processes \citep{Zhou2022LeasttoMostPE, Aggarwal2025L1CH}. 

\begin{figure*}[t]
    \centering
    \includegraphics[width=0.95\linewidth]{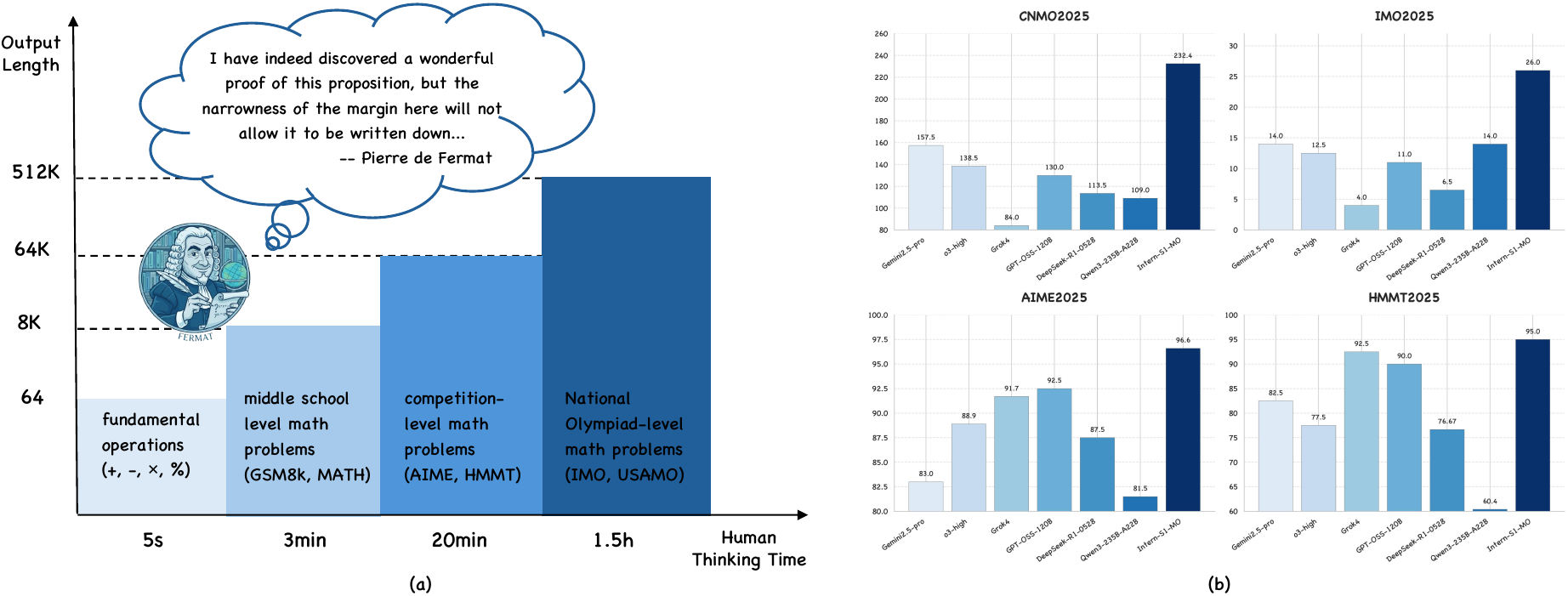}
    \caption{\textbf{The motivation (a) and performance (b) of \methodname{}.}
    More challenging problems generally require larger human thinking time and model reasoning budgets (a), eventually exceeding the capacity of a single context window. \methodname{} enables LRMs to use about 512K tokens to solve a single problem, achieving state-of-the-art performance on challenging mathematical benchmarks (b).}
    \label{fig: teaser}
    \vspace{-1.5em}
\end{figure*}

However, hardware and data limitations have made unlimited scaling of context length infeasible. Currently, state-of-the-art (SOTA) reasoning models typically support a maximum context length of only 64k or 128k tokens \citep{yang2025qwen3, interns1, interns1pro, Comanici2025Gemini2P}, insufficient for ultra-challenging problems such as those in International Mathematical Olympiads (IMO) \footnote{\url{https://imo2025.au}}. Figure \ref{fig: teaser}(a) illustrates the logarithmic growth of the required context length with increasing difficulty of the problem, highlighting the mismatch between the existing capacity limits and practical demands. 
Although resource investment can marginally raise this context ceiling, developing a cost-effective paradigm to meet context requirements is more compelling \citep{Li2025WebThinkerEL, Ke2025ASO, hua2025imitation}.

Some studies have explored multi-round interaction \citep{Motwani2024MALTIR} or parallel decoding \citep{Zhang2024ReSTMCTSLS} to perform long logical deduction in mathematical reasoning. Furthermore, \citet{huang2025gemini} introduced self-reflective with prompt engineering, allowing models to identify flaws in intermediate reasoning steps and refine the results. Nevertheless, these approaches still confine problem-solving to a single reasoning cycle (even with internal iterations) rather than building cumulatively upon prior reasoning trajectories, which limits their capacity to leverage historical explorations for further in-depth deduction \citep{Wang2025ASO}. Alternatively, formal language–based search \citep{Ren2025DeepSeekProverV2AF, chen2025seed, Zhou2025SolvingFM} shows some promise: by maintaining a structured repository to store and reuse intermediate results, they reduce reliance on model context length. However, the proof verification and state traversal demand extensive iterations, leading to high computational and search overhead. Moreover, formal systems require translating informal descriptions into formal logic, introducing additional costs and hindering the interaction between AI and humans.

Proprietary LRMs \citep{openai_imo,gemini_imo} have reported impressive results on the International Mathematical Olympiad 2025 (IMO2025) problems, yet the research community lacks access to their methodologies and models. In this work, we present \methodname{}, a math reasoning agent framework that extends the effective reasoning horizon beyond a single context window through hierarchical decomposition. This strategy closely aligns with human problem-solving patterns. \methodname{} supports repeated exploration through lemma memory management. Specifically, after each single-round reasoning step, the agent compresses its current reasoning history into concise sub-lemmas in a structured memory repository, enabling it to recover historical exploration outcomes in subsequent steps. We furthermore design process verification and revision mechanisms to improve the reliability of the lemma repository. The framework supports configurable reasoning budgets, allowing additional rounds to be allocated to challenging tasks.

To support the bootstrapping and online improvement of \methodname{}, we additionally introduce the OREAL-H framework, enabling the agent to enhance its performance on complex problems with online reinforcement learning (RL). 
Starting from the basic formulation of OREAL \citep{lyu2025exploring}, which is specifically designed for mathematical reasoning, OREAL-H exploits the additional reward signal produced by the outcome process verifier (OPV) that is continuous and accelerates training, and is modified for the Hierarchical Markov Decision Process (MDP) formulation to suit the multi-agent setting of \methodname{}.

Extensive experimental results show that \methodname{} establishes new state-of-the-art results across multiple mathematical reasoning benchmarks. 
As shown in Figure~\ref{fig: teaser}(b), on commonly used inference benchmarks like AIME2025 and HMMT2025, it achieves a 96.6\% and 95\% pass@1 score, respectively, surpassing the current advanced LRMs.
To evaluate the performance of \methodname{} on more difficult, Olympiad-level problems, we test it on the five non-geometry problems of IMO2025, where it obtains 26 out of 35 points. This score is comparable to or higher than that of many silver medalists on the same five-problem subset,\footnote{\url{https://www.imo-official.org/results/individual/year/2025}}.
We also test it on the Chinese National High School Mathematics Olympiad (CNMO2025), which is the preliminary round of the Chinese Mathematics Olympiad (CMO2025)\footnote{\url{https://www.cms.org.cn}}.
CNMO2025 comprises 14 high-school math competition problems (excluding geometry problems), on which our system scores 232.4 out of 260 points.
Additionally, in order to evaluate in a real-world environment, we officially participate in CMO2025 and conduct the test under the same time limit and grading standards as human contestants.
After evaluation by human experts, our system received a score of 102 out of 126, largely exceeding the gold medal threshold of 78 points.


\section{Building Hierarchical Reasoning Agents}
\label{sec 2}

\begin{figure*}[t]
    \centering
    \includegraphics[width=0.9\linewidth]{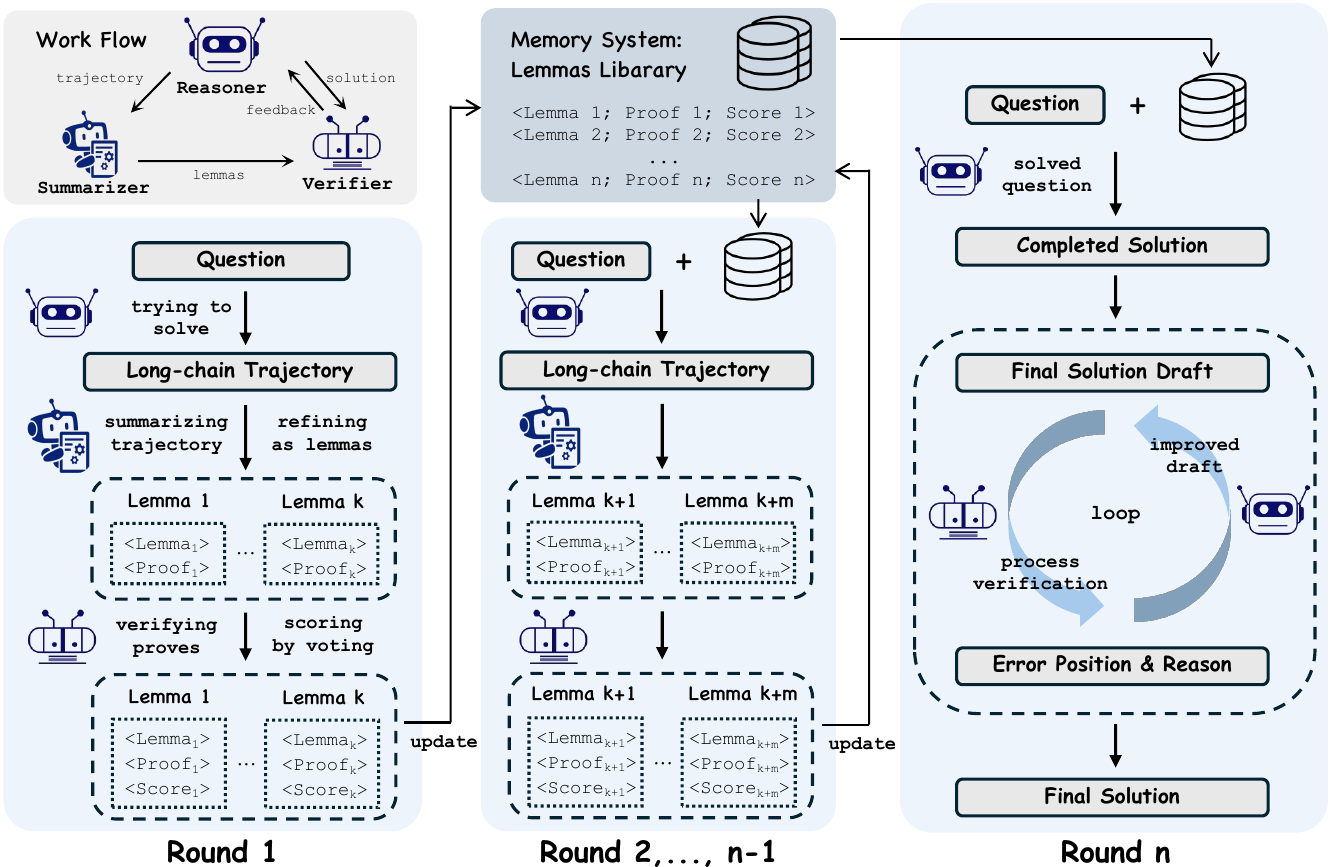}
    \caption{\textbf{The agentic framwork of \methodname{}.} 
    In each reasoning round, the reasoner agent tries to solve the question, and the summarizer agent compresses the current reasoning history into a series of lemmas, which will be added to the memory system after being verified by the verifier agent. 
    Except for the first round, the lemma library will be input into the reasoning agent along with the question.
    In the final round, the solution generated by the reasoner agent undergoes a modification loop, which improves the quality of the solution based on feedback from the verifier agent, until the verification is passed or the maximum number of loop rounds is reached.
    }
    \label{fig: method}
    \vspace{-1.5em}
\end{figure*}

To extend the exploration of reasoning, we designed a hierarchical mathematical reasoning agent tailored for complex competition-level mathematical problems, as shown in Figure~\ref{fig: method}. By enabling recursive subproblem solving, it specifically addresses the aforementioned reasoning limitations constrained by context length. 
In this section, we introduce the main design of the agentic system, including the lemma search, memory maintenance, and verification.
We also give the case study in Appendix \ref{fig:case_study}.

\paragraph{Decomposing Sub-Problems for Lemma Search}
Decomposing complex problems into manageable sub-lemmas is a defining feature of human problem-solving for high-difficulty mathematics, as it breaks long-chain logical reasoning into incremental steps. We first observe that state-of-the-art models already exhibit a degree of reasonable decomposition capability for mathematical problems, though this ability is often undermined by a premature conclusion bias: when reasoning budgets are exhausted, models tend to rush toward incomplete or incorrect final answers instead of acknowledging partial progress. To mitigate this, we refine the model via prompt engineering and targeted training, explicitly enabling it to produce partial deductive progress in single-turn attempts (\eg, deriving intermediate sub-lemmas without forcing a full problem solution). This adjustment aligns the model’s behavior with human iterative reasoning and lays the groundwork for cumulative exploration. The prompts are presented in the Appendix \ref{appendix: prompt}.

\paragraph{Summarizing Exploration for Memory Maintenance}
The model's reasoning processes for complex problems often include redundant exploratory efforts and trial-and-error content~\citep{liu2026thoughtfold}. While this content aids in generating intermediate conclusions, it adds little value to subsequent deductive steps.
Such facts enable us to extract only the essential components that drive progress, specifically, validated intermediate lemmas from each reasoning turn and store them in a structured lemma library. This library encourages the agent to reuse historical conclusions during new exploration rounds, allowing for deeper deductions based on prior lemmas rather than reprocessing similar exploration. 
Notably, summarizing compelling exploration is as complex as the exploration process itself, as it requires distilling and checking the logical validity independently. Therefore, we allocate a dedicated reasoning turn after each exploration step to update the lemma library. This computational cost is necessary to ensure the library remains useful for reasoning.

\paragraph{Verifying Theorems to Mitigate Error Propagation}
Advanced reasoning models can self-reflect, but if they rely on erroneous historical premises, they will expend significant resources trying to validate questionable results. Such a problem is compounded by error propagation, so that a flawed intermediate conclusion can mislead subsequent deductive directions, leading to circular reasoning or invalid proofs~\citep{ji2024anah, gu2024anah, gu2025mask}. Fortunately, the verification of lemmas is comparatively more tractable than that of the complete problem. We address this by integrating a theorem verifier that uses parallel sampling to compute confidence scores for each lemma.
Specifically, for each lemma, we make the theorem verifier perform n parallel verifications, and the proportion of those correctly identified is used as the confidence score. We believe this improves the reliability of theorem verification, avoiding some false positives or false negatives.

\paragraph{Verifying Process for Final Proof Completion}
Verifying the validity of final solutions is crucial for obtaining reliable performance feedback, both in evaluation scenarios and reinforcement learning loops. 
To achieve this, we utilize the process verifier from OPV~\cite{wu2025opv}, whose evaluations demonstrate that their verifier achieves a higher F1-score on ProcessBench~\citep{zheng2024processbench}, surpassing the performance of o1-mini. 
In practice, the verifier serves two main functions: (1) enhancing robustness through test time scaling by aggregating verification results across multiple runs, and (2) providing high-quality feedback signals for iterative revision and reinforcement learning training to further optimize the agent's reasoning precision.

\section{RL training for Evolution of math agents}
\label{sec: 3}

\subsection{Preliminaries}

We model the agentic mathematical reasoning process as a Hierarchical Markov Decision Process, denoted $\mathcal{M} = \langle \mathcal{S}, \mathcal{U}, \mathcal{V}, r, R, \gamma \rangle$, where $\mathcal{S}$ is the state space (problem context + reasoning trace + verification feedback), $\mathcal{U}$ the high-level meta-action space (\eg, "extract lemmas", "invoke verification", "commit answer"), and $\mathcal{V}$ the low-level token vocabulary. The agent alternates between high-level decisions and low-level generation: at each round $t$, it executes a reasoning action $u_t$ with token sequence $\boldsymbol{v}_t = (v_{t,1}, \dots, v_{t,T_t}) \sim \pi^L_\theta(\cdot | s_t)$ to produce a reasoning segment. This output is summarized and verified by an external module, yielding natural language feedback which induces an intermediate proxy reward $r_t \in \mathbb{R}$. Upon termination after several rounds, a sparse final reward $R$ indicates correctness of the solution. The training objective is to maximise expected final reward:
\begin{equation}
    J(\theta, \phi) = \mathbb{E}_{\pi^H_\phi, \pi^L_\theta} \left[ R \right].
\end{equation}
Leveraging the conditional structure of the hierarchical policy, the per-round advantage can be estimated via a high-level critic $V(s_t)$, updated to satisfy:
\begin{equation}
    V(s_t) \leftarrow \mathbb{E} \left[ r_t + \gamma V(s_{t+1}) \right],
\end{equation}
where $s_{t+1}$ is the state after applying $u_t$. The advantage for round $t$ is then $A_t = r_t + \gamma V(s_{t+1}) - V(s_t)$. On a low level, we can then perform an online policy gradient conditioned on this advantage, aggregating token-level log-likelihoods within the round:
\begin{equation}
    \nabla_\theta J = \mathbb{E} \left[ \sum_{t=1}^K A_t \cdot \sum_{\tau=1}^{T_t} \nabla_\theta \log \pi^L_\theta(v_{t,\tau} \,|\, s_t, v_{t,<\tau}) \right].
\end{equation}

\paragraph{Reward Function}  
As mentioned in Section \ref{sec 2}, we employ a Process Verifier (PV) to assess the logical rigor of complex mathematical proofs. Specifically, the PV examines the agent’s final solution and outputs natural language feedback identifying the indices of steps containing logical fallacies. We estimate the PV’s confidence via a multi-round voting mechanism. In particular, for problems amenable to outcome supervision, the final reward $R$ is set to 0 if the final answer is incorrect. We further discuss the role of these supervision signals for RL steps in Section \ref{sec 3.3}.

\subsection{Cloning Success Trajectory for Cold Start}
To prime the agent’s adherence to structured reasoning formats and internalise the iterative agentic workflow, we initialize policies via behavioural cloning on filtered trajectories, retaining only rounds $t$ where the output admits a well-formed lemma summary (\eg, syntactically valid, non-empty, logically segmented). Let $\mathcal{D}_{\text{init}} = \{(s_t, \boldsymbol{v}_t)\}$ denote such transitions. The token-level pretraining objective is:
\begin{equation}
    \mathcal{L}_{\text{RFT}}(\theta) = -\mathbb{E}_{(s_t, \boldsymbol{v}_t) \sim \mathcal{D}_{\text{init}}} \left[ \sum_{\tau=1}^{T_t} \log \pi^L_\theta(v_{t,\tau} \,|\, s_t, v_{t,<\tau}) \right].
\end{equation}
Notably, we continuously augment $\mathcal{D}_{\text{init}}$ with question-answer pairs that are filtered by outcome-based scoring, without previous thinking. We observe that the model exhibits emergent generalization: patterns learned from these simplified trajectories boost agentic solving of the same problems, thereby improving the efficiency of positive trajectory discovery during online RL.

\subsection{OREAL with Conjugate Reward under Process Judgement}
\label{sec 3.3}

We adopt the reinforcement learning framework of OREAL for policy optimization, and introduce two critical adaptations tailored to our Hierarchical MDP setting: (1) credit assignment across high-level reasoning actions is non-trivial due to delayed rewards; (2) the Process Verifier (PV) introduces a continuous, noisy reward signal that deviates from the binary outcome supervision assumed in the RLVR setting. 

\subsubsection{Progress-Conditioned Advantage via Lemma Dependency Graphs}

Existing RLVR training predominantly targets outcome verification (\eg, final answer correctness), which proves insufficient for complex mathematical tasks requiring high process supervision. To align optimization with granular reasoning fidelity, we assign sparse reward signals across reasoning rounds, akin to performing round-level temporal differencing to minimize advantage estimation variance.

To quantify intermediate progress, we construct a lemma dependency graph by aggregating the lemmas generated across multiple rollouts of the same problem. Each node represents an extracted lemma together with its proof and verifier score. Equivalent or highly similar lemmas discovered in different rollouts are merged, while distinct lemmas remain separate nodes. We identify equivalence and dependencies using lexical overlap, semantic similarity, and model-assisted judgments over lemma statements and proofs. If a lemma is judged to use another lemma as a premise or intermediate fact, we add a directed edge from the supporting lemma to the dependent lemma; otherwise, it remains an independent node.

For a trajectory with a positive final reward, lemmas explicitly used in its final proof are designated as terminal success nodes. The trajectory-level reward is injected into these nodes and propagated backward along dependency edges, assigning values to earlier lemmas and to the reasoning rounds that produced them. This graph therefore provides a structured mechanism for attributing delayed proof-level supervision to intermediate mathematical progress. An example is provided in Appendix~\ref{fig: lemmagraph}.

Within this topology, the value of a specific lemma $l$ is not isolated but structurally coupled with the proof's progression. We define the value of a lemma recursively as the expected value of its subsequent derived lemmas, effectively backpropagating the success probability from the final answer to intermediate steps:
\begin{equation}
v(l) = \mathbb{E}_{l' \in \text{Succ}(l)} \left[ v(l') \right],
\end{equation}
where $\text{Succ}(l)$ denotes the set of valid lemmas derived directly from $l$ in the dependency graph. For policy optimization, we anchor credit to rounds that yield verifiable advances. Specifically, for a reasoning round $t$ that generates a set of candidate lemmas $\mathcal{L}_t$, we adopt an optimistic value estimation strategy. We define the state value of round $t$ as the maximum value among its generated candidates, $V(s_t) = \max_{l \in \mathcal{L}_t} v(l)$. The round-level advantage is then computed via the temporal-difference error between the best potential of the current round and the next:
\begin{equation}
A_t = r_t + \gamma \max_{l' \in \mathcal{L}_{t+1}} v(l') - \max_{l \in \mathcal{L}_t} v(l),
\end{equation}
where $r_t$ represents the immediate step reward (\eg, syntactic validity or solving a sub-goal) and $\gamma$ is the discount factor. 

For intermediate rounds yielding no new lemmas ($C_t=0$), the advantage is masked. This formulation drives the gradient estimate with the most promising reasoning path discovered at each step, decoupling optimization intensity from trajectory length and filtering noise from suboptimal branches.

\subsubsection{Conjugate Reward Modeling for Noisy Process Verification}
The lemma graph above determines \emph{where} trajectory-level credit is assigned. We next describe \emph{how} that trajectory-level reward is computed. Because process-verifier votes on complete trajectories are noisy, we convert repeated votes into a calibrated reward before injecting it into the terminal nodes of the graph.

Process Verification (PV) offers valuable insight into the internal logical consistency of a generated solution by subjecting its intermediate steps to multiple stochastic checks. However, unlike final-answer correctness that is deterministic, PV feedback is inherently noisy: a solution passing $k$ out of $n$ verification rounds does not guarantee superior reasoning quality, as passes may arise from lucky sampling or superficial plausibility rather than deep correctness. Directly using the empirical ratio $k/n$ as a reward signal risks amplifying this noise, leading to unstable or misguided policy updates that overfit to verification artifacts rather than genuine mathematical rigor.

To address this, we adopt a Bayesian perspective and model the latent reasoning quality $p \in [0,1]$ as a random variable. We place a uniform prior $p \sim \text{Beta}(1,1)$, encoding no initial assumption about solution validity. After observing $k$ successful verifications in $n$ independent PV trials, the conjugate Beta-Bernoulli update yields the posterior:
\begin{equation}
    p \mid (k,n) \sim \text{Beta}(k+1, n-k+1).
\end{equation}
Instead of using point estimates (\eg, posterior mean), we define the reward as the probability that this solution is strictly better than a canonical completely invalid baseline, one that fails all $n$ checks ($k=0$). Let $p_1 \sim \text{Beta}(k+1, n-k+1)$ represent the quality of the current solution and $p_0 \sim \text{Beta}(1, n+1)$ that of the baseline. The reward is then:
\begin{equation}
\begin{aligned}
    R(k,n) &= \mathbb{P}(p_1 > p_0) \\
    &= \int_0^1 \int_0^1 \mathbb{I}(p_1 > p_0) \cdot f_{\text{Beta}(k+1, n-k+1)}(p_1) \\
    &\quad \cdot f_{\text{Beta}(1, n+1)}(p_0) \, dp_1 dp_0.
\end{aligned}
\end{equation}
This formulation provides a principled, probabilistically calibrated reward that accounts for uncertainty in the verification process. It naturally suppresses spurious signals from low-pass outcomes while preserving strong gradients for high-confidence valid solutions.

In practice, we fix $n=4$, balancing verification cost and signal fidelity. Under this setting, $R(4,4) \approx 0.996$, corresponding to a 99.6\% dominance probability over the completely invalid baseline. We set the reward of that baseline to zero and smoothly interpolate the rewards for intermediate cases ($k=1,2,3$). By grounding the reward in a relative, distributional comparison rather than raw counts, our conjugate reward model denoises PV feedback and aligns policy optimization with latent reasoning quality rather than stochastic verification artifacts. The complete RL training process is presented in Algorithm~\ref{alg:training}.


\section{Experiment}
\label{sec: experiment}

\subsection{Experiment Setup}

We build \methodname{} on Intern-S1~\citep{interns1}, trained on problems from Art of Problem Solving (AoPS) and in-house datasets, verified by CompassVerifier~\cite{liu2025compassverifier} and OPV~\citep{wu2025opv}.
We evaluate on AIME2025~\citep{maaAIME}, HMMT2025~\citep{hmmt2025}, IMO2025, CNMO2025, and CMO2025 (non-geometry parts), against Gemini2.5-pro~\citep{Comanici2025Gemini2P}, o3-high~\citep{o3}, Grok4~\citep{grok4}, GPT-OSS-120B~\citep{openai2025gptoss120bgptoss20bmodel}, DeepSeek-R1-0528~\citep{guo2025deepseek}, and Qwen3-235B-A22B~\citep{yang2025qwen3}.
Full data, implementation, and reasoning-budget details are provided in Appendix~\ref{app: imp_detals}.

\subsection{Overall Results}
\label{subsec:experiment-results}

\begin{table}[t]
\centering
\small
\caption{\textbf{Overall evaluation results for \methodname{} and each baseline.}
Here, the AIME2025 and HMMT2025 scores for the baseline models (first six rows) are from their respective technical reports or corresponding results in Matharena.
For IMO2025, we report the pass@4 score, while the remaining benchmarks report the pass@1 score.
\textbf{Bold} represents the best performance.}
\setlength{\tabcolsep}{3.5pt}
\begin{tabular}{p{1.2in}|cccc}
\toprule
\textbf{Model} & \textbf{HMMT} & \textbf{AIME} & \textbf{CNMO} & \textbf{IMO} \\
\midrule
Gemini2.5-pro          & 82.5  & 83   & 157.5 & 14   \\
o3-high                & 77.5  & 88.9 & 138.5 & 12.5 \\
Grok4                  & 92.5  & 91.7 & 84    & 4    \\
GPT-OSS-120B           & 90    & 92.5 & 130   & 11   \\
DeepSeek-R1-0528       & 76.67 & 87.5 & 113.5 & 6.5  \\
Qwen3-235B-A22B        & 60.4  & 81.5 & 109   & 14   \\
\midrule
\textbf{Intern-S1-MO}  & \textbf{95}    & \textbf{96.6} & \textbf{232.4} & \textbf{26} \\
\bottomrule
\end{tabular}
\label{tab: main_results}
\vspace{-1em}
\end{table}

The quantitative results summarized in Table \ref{tab: main_results} reveal a distinct performance hierarchy, where \methodname{} significantly outperforms the current state-of-the-art baselines. 
On relatively standard competition sets like HMMT2025 and AIME2025, the gap between strong baselines and our method is present but narrower. We hypothesize that performance in these regimes is partially saturated by models capable of pattern matching and heuristic retrieval from pre-training data. For CNMO2025 and IMO2025, where problems demand the construction of novel proof paths, \methodname{} exhibits prominent advantages because it maintains a persistent logical state across multiple reasoning rounds. Unlike single-pass models that have to restart reasoning from the beginning after a failure, our agent can accumulate intermediate progress, such as deriving key inequalities, which simulates the process of step-by-step deduction for human experts solving problems.

Notably, for IMO2025, a score of 26 places \methodname{} within the top percentile of global human competitors, outperforming the national team averages of most participating countries. Preliminary error analysis indicates that the remaining performance gaps mainly stem from problems requiring highly customized transformations or impromptu key constructions, and such problems are difficult to address through systematic search strategies.
Collectively, these findings demonstrate that while parameter scale provides a necessary foundation, the transition from basic competency to proficient mastery in Olympiad-level mathematics requires a structured and verifiable cognitive architecture that enables sustained and multi-step deductive reasoning.

\subsection{Real-world Evaluation: CMO2025}
\label{sec: cmo}

\begin{wraptable}{r}{0.52\textwidth}
\vspace{-1.5em}
\centering
\small
\caption{\textbf{CMO2025 evaluation results.}
Six questions, 21 pts each (126 pts total).}
\setlength{\tabcolsep}{4pt}
\begin{tabular}{c|cccccc}
\toprule
\textbf{Total} & \textbf{P1} & \textbf{P2} & \textbf{P3} & \textbf{P4} & \textbf{P5} & \textbf{P6} \\
\midrule
102 & 21 & 21 & 9 & 21 & 21 & 9 \\
\bottomrule
\end{tabular}
\label{tab:cmo}
\vspace{-1em}
\end{wraptable}

To evaluate \methodname{} in a real-world environment, we officially participate in CMO2025. Similar to human students, our system completed six questions in two days, with a limit of 4.5 hours per day to solve three questions and submit the solutions to the committee immediately. These solutions are scored by human experts using the same standards as those used for human contestants.

We participated in the competition using an extended search budget based on test-time scaling, achieving better results within the given time constraints. For each problem, we performed a 256-shot parallel search over up to 12 rounds. For intermediate lemmas, a lemma verifier provided multiple rounds of 8-shot feedback to help assess and refine their correctness. Upon obtaining candidate solutions, we applied an 8-shot refinement procedure comprising 24 rounds, in which, at each round, the OPV verifier identified informalities or gaps in the proof, which the policy model subsequently revised.

As shown in Table~\ref{tab:cmo}, our system achieves a score of 102 out of 126, exceeding the gold medal threshold of 78 points.
This signifies that \methodname{} not only matches the logical rigor and reasoning ability of top-tier high school math olympiad participants but also transcends the limitations of human problem-solving patterns by independently exploring to discover novel solution methods.

\subsection{Analysis of the Impact of Lemmas}

\begin{wraptable}{r}{0.58\textwidth}
\vspace{-1.5em}
\centering
\small
\caption{\textbf{Analysis on the utility of lemmas.}
``Utility Ratio'' means fraction of generated lemmas explicitly used in the final proof.}
\setlength{\tabcolsep}{4pt}
\begin{tabular}{l|cccc}
\toprule
 & \textbf{HMMT} & \textbf{AIME} & \textbf{CNMO} & \textbf{IMO} \\
\midrule
Utility Ratio & 64.28 & 66.67 & 70.83 & 70.00 \\
\bottomrule
\end{tabular}
\label{tab:ana_lemma}
\vspace{-1em}
\end{wraptable}

In \methodname{}, lemmas serve dual roles: extending effective context length and providing process reward signals for OREAL-H.
To verify that they faithfully compress and transmit logical progress, we measure the utility ratio of generated lemmas.

As shown in Table~\ref{tab:ana_lemma}, we report the number of lemmas generated during multiple rounds of reasoning and the lemmas explicitly used in the final proof.
The results indicate that in high-difficulty benchmarks such as CNMO2025 and IMO2025, approximately 70\% of the explored lemmas directly contribute to the final solution, showing that lemmas perform effective context compression and critical content delivery.
This empirical evidence also supports the validity of the process-conditioned advantage estimation introduced in Section~\ref{sec 3.3}.

To further illustrate how lemmas influence the reasoning process, we provide a case study in Appendix ~\ref{app: case} that demonstrates how lemmas compress long thought chains with a short passage of content. In addition to this qualitative analysis, we also quantitatively analyzed the average length ratio before and after compression, which reaches approximately \textit{64:1}.
Such results ensure that critical historical explorations remain accessible for subsequent reasoning without exhausting the thinking budget.
Appendix~\ref{app: verifier_reliability} further reports a manual evaluation of lemma correctness and verifier reliability, clarifying why verifier scores are retained as soft confidence annotations.

The qualitative cases in Appendix~\ref{app: case} complement these aggregate measurements.
The successful AMO-Bench~\citep{liu-etal-2026-amo} example visualizes how lemmas accumulate and are reused across rounds, while the failure case before RL training illustrates an incorrect bound that motivates verifier-guided refinement and OREAL-H.

\subsection{Ablation Study}
\label{subsec: ablation}

To better understand the contribution of each key component in Intern-S1-MO, we conduct a systematic ablation study.
Due to the limited number of problems in IMO2025 (only five), which brings the volatile results, we compare the evaluation results on HMMT2025, AIME2025, and CNMO2025.

As described in Section~\ref{sec 2} and Section~\ref{sec: 3}, \methodname{} integrates several components, including multi-round reasoning with lemma search and summary, lemma verification, process validation, and an RL framework for training the LRM using the online explored trajectories. 
However, it is crucial to disentangle their individual impacts to validate design choices and assess whether performance gains stem from architectural sophistication or synergistic interactions among modules. 
Therefore, we incrementally build up the full agent system from a simplified baseline, called ``Single-round with Agents'', which means that only one round of inference is performed in the agent system.
Then we progressively add the corresponding component.

\begin{table}[t]
\centering
\small
\caption{\textbf{Ablation study results.}
Here, ``Single-round with Agents'' means that only one round of inference is performed in the agent system, which is the left part of Fig~\ref{fig: method}.
``+ Multi-round Reasoning'' means performing a full multi-round reasoning, but without providing scores for intermediate lemmas and a revised final loop.
``+ Theorem Verifier'' means providing the confidence score for the intermediate lemma, that is, which is the left and middle part of Fig~\ref{fig: method}.
``+ Process Verifier'' means the overall inference workflow.
And ``+ OREAL-H'' means the agents are trained by the RL algorithm introduced in Section~\ref{sec: 3}.
}
\setlength{\tabcolsep}{3.5pt}
\begin{tabular}{p{1.8in}|ccc}
\toprule
\textbf{Model} & \textbf{HMMT} & \textbf{AIME} & \textbf{CNMO} \\
\midrule
Single-round with Agents         & 70.8 & 81.9 & 178.0 \\
\quad + Multi-round Reasoning    & 85.4 & 91.0 & 201.7 \\
\quad + Theorem Verifier         & 86.3 & 93.3 & 203.0 \\
\quad + Process Verifier         & 89.1 & 94.0 & 215.2 \\
\quad + OREAL-H                  & 95.0 & 96.6 & 232.4 \\
\bottomrule
\end{tabular}
\vspace{-1em}
\label{tab: ablation_results}
\end{table}

As shown in Table~\ref{tab: ablation_results}, each agent component brings consistent gains across all benchmarks.
Ultimately, compared with the initial baseline, the full system improves the CNMO2025 score from 178.0 to 232.4 and also produces substantial gains on HMMT2025 and AIME2025.
Appendix~\ref{app: oreal_ablation} further disentangles the lemma-graph credit assignment and conjugate reward modeling within OREAL-H.


\section{Related Work}

\noindent\textbf{Mathematical Reasoning Agents.}
Tree search methods~\citep{Yao2023TreeOT, Zhang2024ReSTMCTSLS} and tool-augmented approaches~\citep{Gou2023ToRAAT, Shao2024DeepSeekMathPT, Huang2025MATHPerturbBL} broaden the inference search space, but often lack sufficient depth to decompose highly complex problems~\citep{sun2025survey, Balunovic2025MathArenaEL}.
More recent structured frameworks~\citep{Yuan2025ReinforceLR, huang2025gemini, zhao2025interngeometry} integrate planning, exploration, and reflection to iteratively refine solutions, outperforming standard CoT on challenging benchmarks.
However, they frequently rely on meticulously designed prompts or human-provided hints, and confine problem-solving to a single reasoning episode without cumulative progress across rounds~\citep{Plaat2024MultiStepRW}.

\noindent\textbf{RL for Math Agents.}
Outcome-reward RL methods~\citep{zhang2024artistimprovinggenerationtextrich, li2025torl, shang2025rstar2agentagenticreasoningtechnical, shen2025semi} yield emergent agentic behaviors such as self-correction and adaptive tool use, and scaling studies~\citep{mai2025agentrlscalinglaw} show increased training effort leads to more sophisticated strategies.
Yet current agents lack cross-episode memory and summarization: decisions remain confined to a fixed reasoning template within isolated episodes.
Approaches like TTRL~\citep{zuo2025ttrl} and Satori~\citep{shen2025satori} introduce basic reflection, while process-aware methods~\citep{kirchner2024proververifiergamesimprovelegibility} provide intermediate supervision via predefined rules—neither supports flexible feedback for natural language proofs, a gap our process verifier addresses.


\section{Conclusion}

This paper aims to address the critical bottleneck in LRMs for complex mathematical reasoning: the inherent limitation of context length, which has hindered progress in solving ultra-challenging tasks such as IMO problems.
To this end, this paper introduces \methodname{}, an LRM-driven multi-agent system that conducts multi-round hierarchical reasoning, which conducts reasoning, summary, and verification at each round.
By maintaining a compact memory in the form of lemmas, Intern-S1-MO can more freely explore the lemma-rich reasoning spaces in multiple reasoning rounds, which significantly extends the 64K constraints of LRMs by about 8 times.
We further propose OREAL-H, an RL framework for training the LRM to simultaneously bootstrap the reasoning ability of the LRM and elevate the overall performance of Intern-S1-MO.
Intern-S1-MO obtains 26 out of 35 points on the five non-geometry problems of IMO2025, a score comparable to that of many silver medalists on the same subset without constituting an official IMO medal result.
It also officially participates in CMO2025 and achieves a score of 102/126 under the judgment of human experts, reaching the gold medal level.
We wish the work paves the way for future research that adopts LRMs for mathematical research.
Moreover, our current experiments focus on mathematical reasoning, where intermediate progress naturally takes the form of lemmas and proofs.
A direction for future work is to study whether the context mechanism can extend to other long-horizon tasks, such as scientific reasoning.

\section*{Acknowledgement}

We thank the anonymous reviewers and area chair for their helpful comments. 
This project is supported by the Shanghai Artificial Intelligence Laboratory. 


\section*{Impact Statements}
This paper presents work whose goal is to advance the field of machine learning. There are many potential societal consequences of our work, none of which we feel must be specifically highlighted here.


\bibliography{reference}
\bibliographystyle{COLM_2026_Template/colm2026_conference}

\newpage
\appendix
\onecolumn


\section{System Prompts for Math Agents}
\label{appendix: prompt}
Our workflow primarily comprises iterative policy lemma search and summarisation, alongside corresponding lemma and final answer verification. Following the final answer verification, the policy model will undergo iterative refinement based on feedback. The prompts for these five actions are presented as follows:

\subsection{Lemma Search}

\lstset{
  basicstyle=\ttfamily\small,
  frame=single,
  backgroundcolor=\color{gray!5},
  breaklines=true,
  postbreak=\mbox{\textcolor{red}{$\hookrightarrow$}\space},
  keepspaces=true,
  columns=fullflexible,
  tabsize=2,
  literate={"}{{\char34}}1,
  showstringspaces=false
}

\begin{lstlisting}[caption={Lemma Search}]
**Objective:**
Your task is to provide a rigorous mathematical proof and solution for the given problem. The problem is expected to be challenging. Your primary goal is to demonstrate a deep and correct understanding of the problem through logical, step-by-step reasoning.

**Guiding Principles:**

1.  **Rigor is Paramount:**
    *   Every step in your proof must be logically sound and clearly justified.
    *   The final answer is secondary to the correctness of the derivation. A correct answer resulting from a flawed or incomplete proof will be considered a failure.

2.  **Embrace Partial Solutions:**
    *   It is understood that a complete solution may not be found in a single attempt.
    *   If you cannot provide a complete solution, you must provide any significant partial results that you can prove with full rigor.
    *   **Do not guess or provide solutions with logical gaps.** Instead, focus on what you *can* prove.
    *   Examples of valuable partial results include:
        *   Proving a key lemma.
        *   Solving one or more cases of a proof by cases.
        *   Establishing a critical property of the mathematical objects involved.
        *   For an optimization problem, proving an upper or lower bound.
    *   Clearly state which parts of the problem you have solved and which remain open. Acknowledging the limits of your solution is a critical part of the task.

3.  **Mathematical Formatting:**
    *   All mathematical variables, expressions, equations, and relations must be formatted using TeX. For example: `Let $G$ be a group and let $H$ be a subgroup of $G$.`

**Output Format:**
Your response MUST be structured into the following sections, in this exact order.

---

**1. Summary**

**a. Verdict:**
*   Begin by stating clearly whether you have found a complete or a partial solution.
*   **For a complete solution:** State the final answer. (\eg, "I have found a complete solution. The answer is...")
*   **For a partial solution:** Clearly state the main rigorous conclusion(s) you have proven (for example: "I have not found a complete solution, but I have rigorously proven the following:"). Your output must strictly follow the Markdown and LaTeX formatting guidelines below:

    - **Format for Proven Lemmas:**
        - All **proven lemmas** and their proofs should be placed together inside a single `\boxed{}` environment.
        - Use `---` horizontal lines to separate different lemmas.
        - Each lemma should begin with `**Lemma X:**`, where `X` is a positive integer.
        - State each lemma concisely and formally, using LaTeX as appropriate.
        - The proof should immediately follow, starting with `**Proof X:**`.
        - Each step of the proof should use an unordered list (`*`), and each step should begin with `**Step Y:**`.

    - **Format for Unproven Lemmas:**
        - All **unproven lemmas** should be placed together in a separate `\boxed{}` environment.
        - Each lemma should begin with `**Lemma X:**`.
        - If all key steps are already provided in the "Provided Lemmas" section or have been fully proven (\ie, **no new unproven lemmas are found**), simply include `**Lemma -1**` in this box.

    - **Example Output Format:**
    ```
    \boxed{
    **lemma n+1**:{lemma n+1}
    **proof n+1**:
    *step 1:{step 1}
    *step 2:{step 2}
    *step 3:{step 3}
    ---
    **lemma n+2**:{lemma n+2}
    **proof n+2**:
    *step 1:{step 1}
    ...
    }
    \boxed{
    **withoutproof**:
    **lemma -1**
    }
    ```

    - After outputting the lemmas, you should end your response immediately without proceeding to the subsequent sections.

**b. Method Sketch:**
*   Provide a high-level, conceptual outline of your logical argument. This should be clear enough for an expert to grasp your approach without reading the full proof.
*   Include:
    *   A narrative of your overall strategy.
    *   The full and precise mathematical statements of any key lemmas or major intermediate results you proved.
    *   A description of any key constructions or case splits that form the backbone of your argument.

**2. Detailed Solution**

*   Present the full, step-by-step mathematical proof of your results.
*   This section should contain *only* the rigorous proof itself, free from any commentary, reflections on your process, or alternative approaches you considered.
*   The level of detail must be sufficient for an expert to verify the correctness of your reasoning without needing to fill in any gaps.
\end{lstlisting}

\subsection{Lemma Summarization}

\begin{lstlisting}[caption={Lemma Summarization}]
You are a top-tier mathematical research assistant, proficient in the logical analysis and argumentation of high-level competitive mathematics.

Your core task is to conduct an in-depth analysis of a solution approach generated by a large language model for problems at the International Mathematical Olympiad (IMO) level, identifying and extracting all key lemmas.

During this analysis, you must rigorously distinguish between propositions **newly proposed** by the model and **universal lemmas** already provided by us. Your final output **shall only contain** those lemmas appearing in the model's solution approach but not provided in the universal lemma repository.

**The input comprises three sections:**
1.  `### Problem ###`: The mathematical problem requiring resolution.
2.  `### Provided Lemmas ###`: A set of known, proven lemmas for reference during problem-solving.
3.  `### Model's Thinking Process ###`: The reasoning process generated by the large language model to solve the problem.

**Your output must adhere to the following principles and format:**

#### **A. Extraction Principles**

1.  **Novelty**: Extract only lemmas first introduced or proven within the `Model's Thinking Process`. Do not include lemmas from the `Provided Lemmas` if the model utilises them.
2.  **Classification**: Extract only new lemmas satisfying the following conditions:
    *   **Proven Lemmas**: Propositions explicitly stated or implicitly utilised within the `model's problem-solving approach`, accompanied by a complete or core proof.

#### **B. Strict Formatting Requirements**

Your output must strictly adhere to the following Markdown and LaTeX formatting.

1.  **Format for Proven Lemmas:**
    *   Each **proven lemma** and its proof must be placed within a separate, non-nested `<lemma>...</lemma>` environment, with the opening and closing tags each occupying a distinct line. The number of `<lemma>...</lemma>` environments must match the number of lemmas extracted in this round. Note that input lemmas may not be presented in this format.
    *   Each lemma must begin on a new line with the text `\n**Lemma X (Lemma X):**`, where `X` is a positive integer numbering. The opening line of this lemma must occupy a complete line. You must strictly adhere to this format. If the lemma has any additional names, annotations, or other descriptions, you may append explanatory text enclosed in brackets after `\n**Lemma X (Lemma X):**`, \eg, `\n**Lemma 2 (Lemma 2):**(Dilworth's Theorem)`.
    *   Subsequently, the remainder of this line must strictly state the content of the lemma. This requires a complete exposition of the lemma extracted from the `Model Problem-Solving Approach`. As the `Model Problem-Solving Approach` frequently introduces entirely new symbols and notations, you must rigorously provide their definitions. Should these definitions involve existing lemmas, you must also specify which particular definitions from those existing lemmas are being referenced.
    *   The statement of the lemma should employ concise, formal mathematical language, utilising LaTeX where appropriate.
    *   This is immediately followed by the proof, commencing on a new line with `**Proof X:**`.
    *   Each step of the proof should begin with an unordered list `*` and be prefixed with `**Step Y:**`. Each step should occupy a separate line. Where an existing lemma is referenced in the proof, you must explicitly state which existing lemma is being referenced and how it is being applied.
    *   The positive integer numbering for each round should be the largest number in the general lemma library incremented by 1. Note that some lemmas in the general lemma library are corrected versions of others. These corrected lemmas share the same numbering as the original lemma, but are marked with the suffix `-fixed`.

Below is a sample input-output pair:

### Problem Statement (Problem) ###
{Problem}

### Provided Lemmas (Provided Lemmas) ###
<lemma>
**Lemma 1 (Lemma 1)**:
**Proof 1 (Proof 1):**:
</lemma>
---...

---
<lemma>
**Lemma n**:
**Proof n:**:
</lemma>

### Model's Thinking Process ###
{Thinking}

---

####`DESIREDOUTPUT:
<lemma>
**Lemma n+1:** lemma n+1
**Proof n+1:**:
* **Step 1:** step 1
* **Step 2:** step 2
* **Step 3:** step 3
</lemma>
<lemma>
**Lemma n+2:** lemma n+2
**Proof n+2:**:
* **Step 1:** step 1
* **Step 2:** step 2
* **Step 3:** step 3
</lemma>
...

\end{lstlisting}

\subsection{Lemma Verify}

\begin{lstlisting}[caption={Lemma Summarization}]

You are a mathematics and logic expert. Your task is to evaluate the correctness of a newly proposed lemma. This lemma relates to the main mathematical problem and may rely on a provided library of existing lemmas.

Your goal is to meticulously check the proof of the new lemma, step by step, to identify the index of the first incorrect step. The index starts at 0 for the first step. If the proof is entirely correct, you should output -1.

Instructions:

- You will be given:
  1. The Main Question: The overarching problem providing context.
  2. Provided Lemmas: A library of existing statements assumed to be correct.
  3. The New Lemma and Its Proof: The student's work to be evaluated, with the proof skeleton broken down into steps.

- A key part of your evaluation is to verify that any use of a lemma from the Provided Lemmas library is correctly applied and that its preconditions are satisfied. The logical inferences within the proof must be sound and either self-evident, derived from the main question's conditions, or justified by one of the provided lemmas.

- You must perform a step-by-step check of the entire solution. Present this analysis as a Detailed Verification Log:
  - Use a numbered list; each item corresponds to a step in the student's proof.
  - For correct steps, provide a brief justification.
  - For steps with errors or gaps, provide a detailed explanation.
  - Do not use a table.

- Finally, at the conclusion of your response, always include a First Error, formatted as `\box{{STEPk}}`, where `k` denotes the index of the first incorrect step. For instance, if step 2 is incorrect, respond with `\box{{STEP2}}`. Should all steps be correct, respond with `\box{{STEP-1}}`.

- The new lemma to verify is guaranteed to possess both a proposition and a proof skeleton. Should any component be missing (thus rendering it an invalid lemma), directly output `FORMAT_ERROR` followed by a description of the observed error. In such instances, no further output is required; omit the `\box{{STEP}}` indicator.

---
### Question ###: 
{Question}

### Historical Lemma Repository (Provided Lemmas) ###
{ProvidedLemmas}

### New Proof to Verify ###
{NewLemmatoVerify}

### Detailed Verification Log and First Error ###:

\end{lstlisting}

\subsection{Final Answer Verify}

\begin{lstlisting}[caption={Final Answer Verify}]


You are a mathematics and educational expert tasked with evaluating the correctness of a student's answer. The student's solution is broken down into steps, and your goal is to identify the index of the first incorrect step. The index starts at zero for the first step. If all steps are correct, you should output -1.

Instructions:
- You will receive a question along with the student's answer, divided into steps. Each step is presented in a separate paragraph.
- You are encouraged to express your internal reasoning within <think>...</think> tags. After presenting your thinking process, you **must** write a summary of your evaluation. Finally, at the end of your response, always include an integer within \\box{{STEP}}. For example, if step 2 is incorrect, respond with \\box{{STEP2}}. If all steps are correct, respond with \\box{{STEP-1}}.
- The student's answer may involve a number of indexed lemmas with their proofs. If you found any of them incorrect, you should report that. For example, if lemma 3 is incorrect, respond with \\box{{LEMMA3}} instead of \\box{{STEP3}}. If all lemmas are correct, you should them detect any incorrect steps. If everything is fine, respond with \\box{{STEP-1}}. Do not count steps inside a lemma. The step index depends on the detailed solution part.
- Some steps may initially appear incorrect but are later corrected in subsequent steps. If a reflection or revision is both accurate and reasonable, the step should be considered correct. If there are multiple reflections, consider only the final one.
- In cases where the problem is ambiguous, consider all possible interpretations and determine if the student's response aligns with any of them.
- Evaluate the entire solution, as some intermediate steps might seem incorrect initially but are rectified later, such as dismissing an extraneous root. Ensure you consider the entire context and, if necessary, review the steps more than once.
- The errors to identify can be very subtle, sometimes hiding in the inexplicit applications of theorems or conditions. So you should actively checking every small logical inferences at a small granularity carefully, either in natural language or in formulas.
- If an error does not affect the overall reasoning, or an gap can be recovered by your effort, you should not report them as incorrect.
- To help you identify the possible errors, every first time you checking a step, you should repeat it in case you missed subtle information. Then you should check its validity by examing its logical inferences within the step/sentences/subsentences one by one.
- Every step should have solid logical basis. Guessing without proof is not allowed.

---
**Question**: {question}
**Reference Answer**: {reference}
**Student's Answer**: {response}
**First Error**:





\end{lstlisting}

\subsection{Self-improve with Verify Feedback}

\begin{lstlisting}[caption={Self-improve with Verify Feedback}]
You are a mathematics and logic expert. Your task is to improve a solution to a math Olympaid problem given a presented solution trial some comments about the solution. 

Your goal is to get a concised, improved solution that solve all errors reasonably pointed out by comments, fill all gaps mentioned by comments and defend other issues that is defendable. Besides, you should compress the complexity of the solution and try your best to make it highschool-level.

Instructions:

- You will be given:
  1. The Main Question: The overall problem providing context.
  2. Provided Solution: The student's work to be evaluated, with a verdict and the complete solution divided into steps.
  3. Previous Comments: Prior attempts to detect specific types of errors. Treat these as helpful guidance rather than authoritative. If they flag something, fix or defend.

- You must present a improved solution with SAME FORMAT. Typically a solution comes with a summary section and a detailed solution section. 

- You are free to decide the idea/approach of your improved solution. You can just fix specific issues, restructure certain parts of the previous answer, or even discard the original solution if considered as unfixable.

- You are adviced to use highschool level of math. If you choose to use university level, then you should treat the readers as smart hihgschool students with no backgrounds, then provide the specific introduction of certain knowledge needed. 

- If the solution to improve contains NO USEFUL INFORMATION (\eg simply admitting its failure). Then you should just return "I have not found a complete solution".
---
### Question
{Question}

### Solution to Improve
{SolutiontoVerify}

### Previous Comments
{PreviousCheckingEfforts}

### Your Improved Solution



\end{lstlisting}

\section{Implementation Details}
\label{app: imp_detals}

\subsection{Experiment Setup Details}

\noindent\textbf{Implementation.}
Our training set contains approximately 5K mathematical problems from 95 sources, including both proof-based and final-answer questions.
The public portion consists of historical competition problems collected from Art of Problem Solving (AoPS)\footnote{\url{https://artofproblemsolving.com/community}}, covering competitions such as IMO, USAMO, AIME, and related olympiad-style contests.
The private portion contains olympiad-level problems from sources such as the Soviet Mathematical Olympiad for high-school students, Chinese national team selection examinations, and Chinese Mathematical Olympiad winter camps.
All training problems were published before 2025, whereas the primary evaluation benchmarks are 2025 competitions.
We additionally perform problem-level deduplication based on normalized problem statements.
These temporal and statement-level filters mitigate, but cannot categorically rule out, potential benchmark contamination.
We generate candidate trajectories using a variant of Intern-S1~\citep{interns1}, then employ the CompassVerifier~\cite{liu2025compassverifier} and OPV~\citep{wu2025opv} as the judger for solution-based and proof-based questions, respectively.
Simultaneously, we select a portion of the challenging problems as RL data, based on the pass rate of Intern-S1 on those data.
Finally, built on Intern-S1~\citep{interns1}, we developed \methodname{}, the multi-agent system solving complex reasoning problems through hierarchical decomposition.

\noindent\textbf{Evaluation.}
We use well-established mathematical datasets for evaluation: AIME2025~\citep{maaAIME}, HMMT2025 Feb~\citep{hmmt2025}, IMO2025, CNMO2025, and CMO2025.
For CNMO2025 and IMO2025, we only evaluate the non-geometric parts.
Referring to the approach of MathArena~\citep{Balunovic2025MathArenaEL}, we build an evaluation system that scores answers based on fine-grained grading points (details in Appendix~\ref{appendix: eval}), employed for IMO2025 and CNMO2025.
For each sample, we perform 16 independent rollouts and use the unbiased \texttt{pass@1}~\citep{chen2021evaluating} as the metric, except for IMO2025, which uses \texttt{pass@4}.
For CMO2025, we participate under the official contest time window and grading protocol, but use parallel search and therefore do not claim a compute-equivalent comparison with human contestants.

\noindent\textbf{Baselines.}
We compare against Gemini2.5-pro~\citep{Comanici2025Gemini2P}, o3-high~\citep{o3}, Grok4~\citep{grok4}, GPT-OSS-120B~\citep{openai2025gptoss120bgptoss20bmodel}, DeepSeek-R1-0528~\citep{guo2025deepseek}, and Qwen3-235B-A22B~\citep{yang2025qwen3}.
For AIME2025 and HMMT2025, baseline scores are from their respective technical reports or MathArena.

Our agentic system is a scalable framework that allows for custom inference budgets based on problem difficulty.
Theoretically, a higher inference budget leads to better performance, which aligns with the core logic of the TTS (test time scaling) strategy~\citep{muennighoff2025s1}.

To control evaluation costs, the default inference budget uses at most 8 reasoner--summarizer rounds, 4 parallel theorem-verifier checks for each lemma, and at most 8 final revision rounds based on process-verifier feedback.
The reasoner and summarizer each have a maximum output length of 64K tokens.
Across the benchmarks reported in Table~\ref{tab: main_results}, the system uses 4.48 reasoning rounds per problem on average.
RL rollouts follow the same default reasoning-round and verification budgets.
For the CMO2025 evaluation, we use a larger contest-time budget: 256-way parallel search for each problem, up to 12 reasoning rounds, 8-shot lemma-verifier feedback, and an 8-shot final refinement procedure with up to 24 rounds.

\subsection{Hyperparameters Details}

During training iterations, each batch consists of 64 questions, with 16 rollouts per question. The max length of each rollout trajectory is set to 65536 tokens. Then the correctness of each response is averaged to calculate the pass rate, and questions with an overall pass rate of 0 or 1 are discarded.

For optimization, the policy model is trained with a learning rate of $5e{-7}$. Both models employ a cosine annealing learning rate schedule, decaying to $1/5$ of the initial learning rate over time. We optimize both models using the AdamW optimizer. The KL coefficient $\beta$ is set to 0.01.

\subsection{Token-Budget-Matched Parallel Sampling}
\label{app: token_matched}

Results for several frontier baselines are taken from public reports or MathArena, whose hidden reasoning budgets and inference pipelines are not fully controllable.
We therefore conduct an additional comparison with the open-weight GPT-OSS-120B baseline on IMO2025 under the same grading protocol.
One \methodname{} inference run consumes 391.7K generated tokens per problem on average, whereas one GPT-OSS-120B rollout consumes 55.9K tokens.
We consequently sample 7 GPT-OSS-120B rollouts per problem and select the highest-scoring solution, yielding a nearly matched aggregate token budget.

\begin{table}[h]
\centering
\small
\caption{\textbf{Token-budget-matched comparison on IMO2025.}
All systems are evaluated with the same grading protocol on the five non-geometry problems.
The matched GPT-OSS-120B setting uses 7-way parallel sampling.}
\label{tab: token_matched}
\setlength{\tabcolsep}{8pt}
\begin{tabular}{lcc}
\toprule
\textbf{Model} & \textbf{Avg. tokens/problem} & \textbf{IMO2025 score} \\
\midrule
GPT-OSS-120B & 55.9K & 11 \\
GPT-OSS-120B, 7-way sampling & 391.3K & 18 \\
\methodname{} & 391.7K & 26 \\
\bottomrule
\end{tabular}
\end{table}

Parallel sampling improves GPT-OSS-120B from 11 to 18, but remains below \methodname{} under a comparable generated-token budget.
This comparison suggests that naive parallel sampling alone does not account for the full improvement.
It is not a strict compute-matched comparison, however, because hardware utilization, wall-clock cost, and implementation details differ between systems.

\subsection{Lemma Verifier Reliability}
\label{app: verifier_reliability}

We randomly sample 50 generated lemmas and manually annotate whether each statement and proof is mathematically valid under the problem context and preceding lemma library.
Each lemma receives 4 parallel theorem-verifier checks.
For this analysis, a lemma is classified as verifier-accepted when its score is greater than 0.25 and verifier-rejected otherwise.

\begin{table}[h]
\centering
\small
\caption{\textbf{Lemma-verifier reliability on 50 manually annotated lemmas.}}
\label{tab: verifier_reliability}
\setlength{\tabcolsep}{8pt}
\begin{tabular}{lccc}
\toprule
 & \textbf{Human correct} & \textbf{Human incorrect} & \textbf{Total} \\
\midrule
Verifier accepted & 27 & 2 & 29 \\
Verifier rejected & 6 & 15 & 21 \\
\midrule
Total & 33 & 17 & 50 \\
\bottomrule
\end{tabular}
\end{table}

The verifier obtains 93.1\% precision, 81.8\% recall, 87.1\% F1, and 84.0\% accuracy.
We therefore use verifier scores as soft reliability annotations rather than hard filters.
Low-confidence lemmas are marked as uncertain but retained because false negatives may still contain useful search directions; later reasoning may reprove or correct them before they contribute to a verified final proof.

\subsection{OREAL-H Component Ablation}
\label{app: oreal_ablation}

To separate the effects of reward modeling and credit assignment, we evaluate two intermediate OREAL-H variants.
The first retains conjugate reward modeling but removes graph-based credit assignment, so progress rounds in the same trajectory share the trajectory-level advantage.
The second retains the lemma dependency graph but replaces the conjugate reward with the raw process-verifier pass ratio $k/n$, which is injected into terminal graph nodes and propagated to earlier lemmas and rounds.

\begin{table}[h]
\centering
\small
\caption{\textbf{Component ablation of OREAL-H.}
The lemma dependency graph and conjugate reward modeling provide complementary improvements.}
\label{tab: oreal_ablation}
\setlength{\tabcolsep}{8pt}
\begin{tabular}{lccc}
\toprule
\textbf{Setting} & \textbf{HMMT} & \textbf{AIME} & \textbf{CNMO} \\
\midrule
w/o OREAL-H & 89.1 & 94.0 & 215.2 \\
OREAL-H w/o lemma dependency graph & 91.35 & 94.38 & 219.0 \\
OREAL-H w/o conjugate reward modeling & 93.18 & 95.73 & 226.4 \\
Full OREAL-H & 95.0 & 96.6 & 232.4 \\
\bottomrule
\end{tabular}
\end{table}

Both components improve performance across all three benchmarks.
Removing the lemma dependency graph causes the larger degradation, particularly on CNMO2025, indicating the importance of propagating delayed rewards to intermediate mathematical progress.
Replacing the conjugate reward with raw verifier pass ratios also consistently reduces performance, supporting its role in calibrating noisy process-verifier feedback.
Full OREAL-H performs best, showing that graph-based credit assignment and verifier-noise calibration are complementary.

\section{Grading Details}
\label{appendix: eval}

Automated evaluation of complex mathematical proofs presents substantial challenges. LLMs often exhibit excessive sensitivity to lexical phrasing while occasionally overlooking missing logical reasoning. To bridge the gap between automated evaluation pipelines and human experts, we designed a fine-grained grading scheme tailored to the nature of the problem.

\paragraph{Calculation-Centric Evaluation (HMMT, AIME)}
For datasets primarily focused on final answers, such as HMMT and AIME, we only employ \textbf{Final Answer Accuracy} as the sole metric. A response is awarded full score if and only if the extracted final answer matches the ground truth exactly; otherwise, it receives zero.

\paragraph{Proof-Oriented Evaluation (CNMO, IMO)}
For Olympiad-level proof problems (\eg, CNMO and IMO), we adopt a rubric-based scoring logic inspired by MathArena \cite{Balunovic2025MathArenaEL}, with critical modifications to ensure rigor.

The key difference here is their grading schemes often list only the necessary sub-propositions, lacking explicit constraints on the derivation of conclusions. When used with LLM-based judges, this ambiguity frequently leads to significant false positives. To rectify this, we augmented the grading scheme by explicitly coupling sub-propositions with their corresponding conclusion requirements.
A representative example of our revised grading scheme is shown in FIgure. \ref{lst:grading_scheme}.

\begin{footnotesize}
\begin{lstlisting}
[
	{
		"desc": "1 point should be given for rigorously describing a construction for $n$=3. Should prove that $k=2$ is impossible, and that $k=0,1,3$ is possible.",
		"points": 1,
		"title": "Describing a construction for $k=0,1,3$ for $n=3$"
	},
	{
		"desc": "1 point should be given for just finding the answer $k=0,1,3$ is valid for all $n$. Providing a specific construction is sufficient to earn points.",
		"points": 1,
		"title": "Reaching the answer $k=0,1,3$ for all $n$"
	},
	{
		"desc": "Stating and proving that the perimeter sides ($x=1, y=1, x+y=n+1$) contain a total of $3n-3$ points.",
		"points": 1,
		"title": "Making an statement about the boundary points"
	},
	{
		"desc": "Arguing that any sunny line can cover at most two points on the perimeter sides, so for $n>3$, there must be at least one non-sunny line covering a complete boundary line.",
		"points": 2,
		"title": "Proving the existence of a non-sunny line covering a complete boundary line"
	},
	{
		"desc": "Stating and proving that if a non-sunny line contains one of the 3 perimeter sides ($x=1, y=1, x+y=n+1$), the problem can be to reduce for $n-1$ without changing the answer.",
		"points": 1,
		"title": "Reducing the problem from $n$ to $n-1$ given a boundary line"
	},
	{
		"desc": "Finishing by summarizing the final answer that for any $n$, the possible values of $k$ are 0, 1, and 3.",
		"points": 1,
		"title": "Finishing"
	}
]
\end{lstlisting}
\textbf{Example of the Refined Grading Scheme for IMO2025 P1.} This JSON structure outlines the specific proof obligations, point allocation, and partial credit policies used to guide the LLM judge.
\end{footnotesize}
\label{lst:grading_scheme}

To mitigate the inherent stochasticity of LLM judges, we implement an ensemble evaluation protocol. Each generated solution is evaluated in parallel across $N=8$ independent runs. For grading points worth more than 1 point, the model is awarded partial credit if it provides a valid partial proof. The final score for a solution is calculated as the arithmetic mean of the total scores obtained across the eight evaluation runs.

\newpage

\section{OREAL-H Algorithm}

The complete RL training procedure is described in Algorithm~\ref{alg:training}.

\begin{algorithm}[h]
\caption{Evolution of Hierarchical Math Agents via Lemma Dependency Graphs}
\label{alg:training}
\begin{algorithmic}[1]
\REQUIRE Policy $\pi_\theta$, Verifier (PV) $V_\phi$, Dataset $\mathcal{D}$, Hyperparameters $\gamma, \alpha$
\REQUIRE Conjugate Reward function $R_{\text{conj}}(k,n)$ (Eq. 7)

\WHILE{not converged}
    \STATE Initialize batch buffer $\mathcal{B} \leftarrow \emptyset$
    \FOR{each problem $q \in \text{Batch}(\mathcal{D})$}
        \STATE \textbf{\# Step 1: Multi-round Rollout}
        \STATE Sample $K$ trajectories $\{\tau^{(i)}\}_{i=1}^K$ using $\pi_\theta(\cdot | q)$
        
        \STATE \textbf{\# Step 2: Process Verification}
        \STATE Evaluate each trajectory $\tau^{(i)}$ via $V_\phi$
        \STATE Compute final reward $R^{(i)} \leftarrow R_{\text{conj}}(k, n)$
        
        \STATE \textbf{\# Step 3: Construct Lemma Dependency Graph}
        \STATE Initialize graph $\mathcal{G} = (\mathcal{V}, \mathcal{E})$ with lemmas from all $\{\tau^{(i)}\}$
        \STATE Identify terminal nodes where $R^{(i)} \neq 0$
        \STATE Backpropagate values recursively to compute lemma values:
        \STATE \quad $v(l) \leftarrow \mathbb{E}_{l' \in \text{Succ}(l)} [v(l')]$ \# Eq. (4)
        
        \STATE \textbf{\# Step 4: Compute Progress-Conditioned Advantage}
        \FOR{each trajectory $\tau^{(i)}$ and step $t=1 \dots T$}
            \IF{$C_t^{(i)} = 1$} 
                \STATE Estimate state value: $V(s_t) \leftarrow \max_{l \in \mathcal{L}_t} v(l)$
                \STATE Compute next-state value: $V(s_{t+1}) \leftarrow \max_{l' \in \mathcal{L}_{t+1}} v(l')$
                \STATE Calculate advantage $A_t^{(i)} \leftarrow r_t + \gamma V(s_{t+1}) - V(s_t)$ \# Eq. (5)
            \ELSE
                \STATE $A_t^{(i)} \leftarrow 0$ \# Mask non-progress rounds
            \ENDIF
            \STATE Add $(s_t, u_t, A_t^{(i)})$ to $\mathcal{B}$
        \ENDFOR
    \ENDFOR
    
    \STATE Optimization via OREAL Loss \citep{lyu2025exploring}
\ENDWHILE
\end{algorithmic}
\end{algorithm}

\newpage

\section{Lemma Graph}
\label{fig: lemmagraph}

\begin{figure}[htbp]
    \centering
    \includegraphics[width=0.95\linewidth]{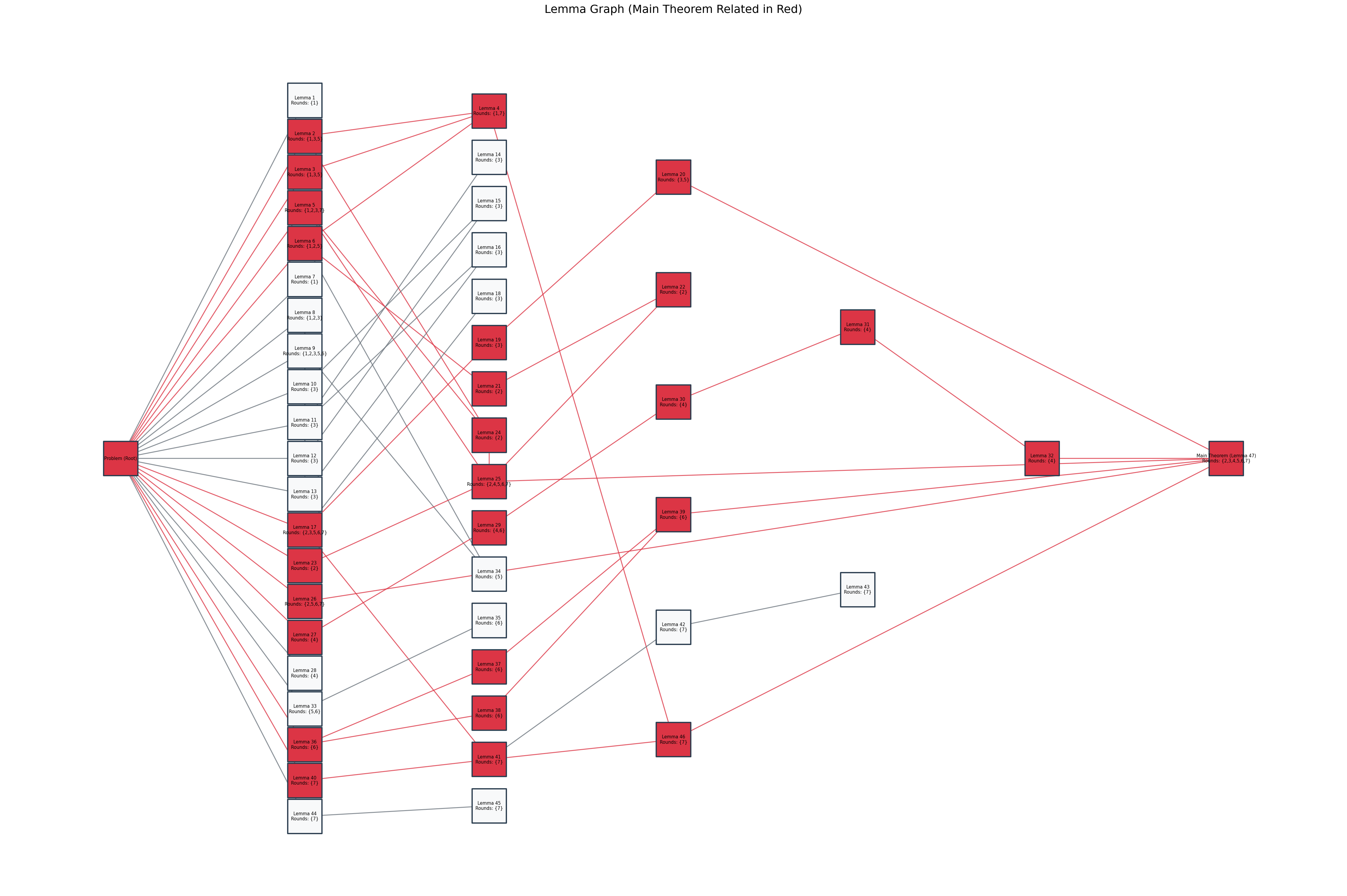}
    \caption{\textbf{Example of a lemma graph.} 
    Red nodes mark all lemmas contributing to the final conclusion, with the numbers below indicating the number of rollout in which each lemma appeared.
    }
    \vspace{-1em}
\end{figure}

\newpage
\section{Case Example}
\label{app: case}

We present a detailed reasoning trace for Problem 1 from AMO-Bench~\citep{liu-etal-2026-amo}.

\centering
\begin{tcolorbox}[
    enhanced,
    breakable,
    title={\textbf{Case Study: AMO-Bench P1}},
    colback=white,
    colframe=gray!25!black,
    coltitle=white,
    fonttitle=\bfseries\large,
    attach boxed title to top left={xshift=5mm, yshift=-2mm},
    boxed title style={colback=blue!60!black, sharp corners},
    drop shadow
]

\begin{tcolorbox}[
    enhanced,
    breakable,
    blanker,
    borderline west={3pt}{0pt}{blue!60!black},
    left=10pt,
    bottom=10pt
]
    \textbf{Problem Statement:} Divide each side of an equilateral triangle $ABC$ into four equal parts. Through each division point, draw lines parallel to the other two sides, defining 15 lattice points. Find the smallest integer $n$ such that, no matter which $n$ points are chosen, there always exist three points that form an isosceles triangle.
\end{tcolorbox}
\end{tcolorbox}

\vspace{0.2em}
\textbf{\small $\triangleright$ Reasoning Process \& Lemma Accumulation}
\vspace{0.2em}

\begin{tcolorbox}[
    breakable, colback=blue!5!white, colframe=blue!20!white, arc=2mm, title={\small \textbf{Round 1}}, coltitle=black, fonttitle=\bfseries\small]
    \small
    \noindent \textbf{Lemma 1:} \\
The 15 lattice points obtained by dividing each side of an equilateral triangle $ABC$ into four equal parts can be uniquely represented by integer pairs $(i,j)$, where
\[
i \ge 0, \qquad j \ge 0, \qquad i+j \le 4.
\]
Let $P=(i_{1},j_{1})$ and $Q=(i_{2},j_{2})$ be any two points. Their squared Euclidean distance satisfies
\[
|PQ|^{2}= \Delta i^{2}+\Delta i\,\Delta j+\Delta j^{2},
\qquad\text{where }\Delta i=i_{1}-i_{2},\ \Delta j=j_{1}-j_{2}.
\]

\noindent \textbf{Proof 1:}
\begin{itemize}
    \item \textbf{Step 1:} Establish a coordinate system with vertex $A$ as the origin, and basis vectors $\mathbf{u}=(1,0)$ and $\mathbf{v}=\bigl(\tfrac12,\tfrac{\sqrt3}{2}\bigr)$.
    \item \textbf{Step 2:} The Cartesian coordinates of any point $(i,j)$ are $\;i\mathbf{u}+j\mathbf{v}=(i+\tfrac{j}{2},\tfrac{\sqrt3}{2}j)$.
    \item \textbf{Step 3:} For two points $(i_{1},j_{1})$ and $(i_{2},j_{2})$, calculate the difference vector and apply the Euclidean distance formula. Simplification yields
    \[
    |PQ|^{2}=(\Delta i+\tfrac{\Delta j}{2})^{2}+\bigl(\tfrac{\sqrt3}{2}\Delta j\bigr)^{2}
    =\Delta i^{2}+\Delta i\,\Delta j+\Delta j^{2}.
    \]
\end{itemize}
(The confidence score is 0.75.)

\vspace{1em}

\noindent \textbf{Lemma 2:} \\
Let $P=(a,0)$ and $Q=(b,0)$ be two lattice points on the same side $AB$ (where $0 \le a < b \le 4$). Then, any lattice point $X=(i,j)$ (satisfying $i,j \ge 0,\ i+j \le 4$) satisfies
\[
|XP|=|XQ| \quad \Longleftrightarrow \quad 2i+j=a+b.
\]

\noindent \textbf{Proof 2:}
\begin{itemize}
    \item \textbf{Step 1:} Based on Lemma 1, write
    $|XP|^{2}=(i-a)^{2}+(i-a)j+j^{2}$ and $|XQ|^{2}=(i-b)^{2}+(i-b)j+j^{2}$.
    \item \textbf{Step 2:} Subtract the two equations and cancel common terms to obtain
    $(i-a)^{2}-(i-b)^{2}+(i-a)j-(i-b)j=0$.
    \item \textbf{Step 3:} Simplify to $(b-a)(2i+j-a-b)=0$. Since $a \neq b$, we obtain the required linear relation $2i+j=a+b$.
\end{itemize}
(The confidence score is 0.0.)

\vspace{1em}

......

\noindent \textbf{Lemma 10:} \\
Among all 15 lattice points, the cardinality of the largest point set containing no isosceles triangles is 5, and only sets of the form ``all 5 lattice points of one side'' can achieve this upper bound.

\noindent \textbf{Proof 10:}
\begin{itemize}
    \item \textbf{Step 1:} By Lemma 8, taking all 5 lattice points of any single side yields an isosceles-free set of size 5.
    \item \textbf{Step 2:} If we attempt to construct an isosceles-free set of size 6, there are two cases:
    \begin{itemize}
        \item It contains internal lattice points: This contradicts Lemma 6.
        \item It contains only boundary lattice points: This contradicts Lemma 9.
    \end{itemize}
    Thus, no isosceles-free set of size 6 exists.
    \item \textbf{Step 3:} Therefore, the maximum cardinality is 5, and the only maximal construction is the set of all lattice points on a single side.
\end{itemize}
(The confidence score is 0.0.)
    
\end{tcolorbox}

\begin{tcolorbox}[
    breakable, colback=blue!5!white, colframe=blue!20!white, arc=2mm, title={\small \textbf{Round 2}}, coltitle=black, fonttitle=\bfseries\small]
    \small
    Can not summary any new lemmas.
    
\end{tcolorbox}

\begin{tcolorbox}[
    enhanced,
    breakable, colback=blue!5!white, colframe=blue!20!white, arc=2mm, title={\small \textbf{Round 3}}, coltitle=black, fonttitle=\bfseries\small]
    \small
\noindent \textbf{Lemma 11:} \\
In the set of triangular lattice points, for any two points $P$ and $Q$ chosen from two different sides, there exists a unique third point $R$ (located on the third side or in the interior) such that $|RP|=|RQ|$.

\noindent \textbf{Proof 11:}
\begin{itemize}
    \item \textbf{Step 1:} Let $P$ and $Q$ be located on sides $AB$ and $AC$, respectively. Denote their coordinates as $P=(a,0)$ and $Q=(0,b)$, where $0 \le a,b \le 4$.
    \item \textbf{Step 2:} According to Lemma 1, calculate the difference in squared distance between the two points:
    \[
    |RX|^{2}-|RY|^{2}= (i-a)^{2}+(i-a)j+j^{2}-(i)^{2}-(i)(j-b)-(j-b)^{2},
    \]
    and simplify to obtain the equidistance condition:
    \[
    2i+j=a+b.
    \]
    \item \textbf{Step 3:} Under the constraints $i,j \ge 0,\ i+j \le 4$, the equation $2i+j=a+b$ has a unique integer solution $(i_{0},j_{0})$ within the triangle. If this solution satisfies $j_{0}>0$, then the point $R=(i_{0},j_{0})$ is located on side $AC$; if $j_{0}=0$, then $R$ is located on side $BC$. In any case, $R$ and $P, Q$ form an isosceles triangle.
    \item \textbf{Step 4:} It can similarly be proven that if $P, Q$ are located on any two different sides, a unique point $R$ satisfying the equidistance condition can be found.
    \item \textbf{Step 5:} In summary, for any two points on different sides, there must exist a unique third point equidistant to them, completing the proof.
\end{itemize}
(The confidence score is 0.0.)

\vspace{1em}

\noindent \textbf{Lemma 12:} \\
In any set consisting only of points on the three sides, if the cardinality of the set is 6, then the distribution of points across the sides must be such that no more than two points lie on any single side; otherwise, an isosceles triangle must exist.

\noindent \textbf{Proof 12:}
\begin{itemize}
    \item \textbf{Step 1:} Let set $S$ contain $x$ points from the sides, where $x \ge 3$. If any side contains at least three points, select any two points $P, Q$ on that side. By Lemma 4, there exists a third point $R$ (not on that side) such that $|RP|=|RQ|$, thus forming an isosceles triangle.
    \item \textbf{Step 2:} Therefore, to avoid an isosceles triangle, the number of points on any single side must be at most 2.
    \item \textbf{Step 3:} Assume the cardinality of $S$ is 6. The distribution across the three sides must be $(2,2,2)$ or $(2,3,1)$ (partitioned by the three sides). If $(2,3,1)$ occurs, the side containing three points must generate an isosceles triangle, a contradiction. Thus, the only possible distribution is $(2,2,2)$.
    \item \textbf{Step 4:} For the $(2,2,2)$ configuration, let two points be located on different sides, denoted $(a,0)$ and $(0,b)$. By Lemma 11, a unique third point $R$ satisfying $|RP|=|RQ|$ can be determined. This point must lie on the third side, and thus must be in the set $S$, resulting in an isosceles triangle.
    \item \textbf{Step 5:} Therefore, any set of size 6 consisting only of boundary points must contain an isosceles triangle, completing the proof.
\end{itemize}
(The confidence score is 0.0.)

\end{tcolorbox}

\begin{tcolorbox}[
    breakable, colback=blue!5!white, colframe=blue!20!white, arc=2mm, title={\small \textbf{Round 4}}, coltitle=black, fonttitle=\bfseries\small]
    \small
\noindent \textbf{Lemma 13:} \\
In any set of 6 points consisting only of points on the three sides (\ie, containing no internal points), the arrangement wherein at most two points are on any single side cannot avoid an isosceles triangle. In other words, if a point set $S\subseteq\{(i,j)\mid i\ge0,\ j\ge0,\ i+j\le4\}$ contains $6$ points and no internal points, then there must exist three points $P,Q,R\in S$ such that $\lvert RP\rvert=\lvert RQ\rvert$, thereby forming an isosceles triangle.

\noindent \textbf{Proof 13:}
\begin{itemize}
    \item \textbf{Step 1:} Let $x,y,z$ be the number of points in $S$ on sides $AB, AC, BC$, respectively, such that $x+y+z=6$. If the number of points on any side is $\ge 3$, select any two points $P,Q$ on that side. By Lemma 4 (two points on the same side must have a third equidistant point), there exists a point $R \neq P,Q$ (located on another side or internally) such that $\lvert RP\rvert=\lvert RQ\rvert$. If $R\in S$, an isosceles triangle is formed. If $R\notin S$, then $S$ contains at most all 5 points of that side, and since $x\ge3$ and $x\le5$, the possible 6-point distributions must be unbalanced, such as $(3,2,1)$, $(3,1,2)$, $(4,1,1)$, $(5,1,0)$, etc. Upon checking these distributions, the equidistant point $R$ corresponding to $P,Q$ can always be found on another side, and since that side has at least one point selected into $S$, $R$ must be in $S$, resulting in an isosceles triangle.
    \item \textbf{Step 2:} Therefore, to avoid an isosceles triangle, the number of points on each side must be $\le 2$. Since $x+y+z=6$, the only possible distribution is $(2,2,2)$.
    \item \textbf{Step 3:} Assume the two points on side $AB$ are $P=(a,0), Q=(b,0)$ ($a<b$). The equidistant point satisfies the line equation $2i+j=a+b$ (Lemma 2). Consulting the table (from Lemma 4), this equation has at least one solution $R$ within the triangle, and $R$ must be on another side (if $a+b$ is odd, $R$ is not on $AB$). Similarly, the point pair on side $AC$ yields the equation $i+2j=c+d$, and the pair on side $BC$ yields $i-j=e+f-4$ (a corollary of Lemma 4).
    \item \textbf{Step 4:} Since there are exactly two points on each side, there must exist a pair of points such that their equidistant point $R$ falls onto a selected point on another side. Specifically, if $a+b$ is odd, the unique integer solution to $2i+j=a+b$ lies on $BC$, $R=(\frac{a+b-1}{2},\frac{a+b+1}{2})$, and since $BC$ contains exactly two points, one of them must be $R$. If $a+b$ is even, the midpoint $(\frac{a+b}{2},0)$ is on $AB$, but since only two points were selected on $AB$, $\frac{a+b}{2}$ must not equal $a$ or $b$. The equidistant point must then lie on $BC$ or $AC$, and similarly, it must fall into the already selected point set. Consequently, an isosceles triangle must appear.
    \item \textbf{Step 5:} In summary, any set of 6 points containing no internal points must include an isosceles triangle.
\end{itemize}
(The confidence score is 0.0.)
    
\end{tcolorbox}

\vspace{0.5em}
\begin{tcolorbox}[
    enhanced,
    breakable, colback=yellow!10!white, colframe=orange!50!yellow, boxrule=1pt, arc=2mm]
    \textbf{Final Answer:} 
    \small

\textbf{The Triangular Lattice}

We place the equilateral triangle $ABC$ of side length $4$ such that its vertices are:
\[
A=(0,0),\qquad B=(4,0),\qquad C=\Bigl(2,2\sqrt{3}\Bigr).
\]
Let the basis vectors for the lattice be:
\[
\mathbf{u}=(1,0),\qquad 
\mathbf{v}=\Bigl(\tfrac{1}{2},\tfrac{\sqrt{3}}{2}\Bigr).
\]
Every point obtained by the construction is a lattice point, representable by integer coordinates $(i,j)$:
\[
(i,j) = i\mathbf{u}+j\mathbf{v} = \Bigl(i+\tfrac{j}{2},\tfrac{\sqrt{3}}{2}j\Bigr),
\qquad i,j\in\mathbb Z_{\ge0},\;i+j\le4.
\tag{1}
\]
Thus, the 15 points constitute the set:
\[
\mathcal L=\{(i,j)\mid i,j\ge0,\;i+j\le4\}.
\]
For two points $P=(i_1,j_1)$ and $Q=(i_2,j_2)$, the squared Euclidean distance is obtained by substituting coordinates from (1) and simplifying:
\[
|PQ|^{2}=(i_1-i_2)^{2}+(i_1-i_2)(j_1-j_2)+(j_1-j_2)^{2}. \tag{2}
\]

\textbf{Interior Points Force an Isosceles Triangle}

The interior points of $\mathcal L$ (those not on the boundary) are:
\[
I_{1}=(1,1),\qquad I_{2}=(1,2),\qquad I_{3}=(2,1).
\]
Using the distance formula (2), one verifies that any interior point possesses **exactly four distinct squared distances** to the remaining fourteen lattice points:
\[
\begin{array}{c|c}
\text{Point}&\text{Possible }|P-Q|^{2}\\\hline
(1,1)&1,\;3,\;4,\;7\\
(1,2)&1,\;3,\;4,\;7\\
(2,1)&1,\;3,\;4,\;7
\end{array}
\]
Since an interior point can be at at most four different distances from all other points of $\mathcal L$, consider a set $S\subseteq\mathcal L$ that contains an interior point $I$, with $|S|=6$. The five other points of $S$ yield five distances from $I$. By the **Pigeonhole Principle**, two of these distances must be equal. These two points, together with $I$, form an isosceles triangle. Consequently, **any six-point set that contains an interior point necessarily contains an isosceles triangle**.

\textbf{Six Boundary Points Also Contain an Isosceles Triangle}

We now restrict our attention to the case where all six chosen points lie on the boundary of the triangle. The boundary points lie on the three sides:
\[
\begin{aligned}
AB&:\;(i,0),\;i=0,1,2,3,4,\\
AC&:\;(0,j),\;j=0,1,2,3,4,\\
BC&:\;(i,4-i),\;i=0,1,2,3,4,
\end{aligned}
\tag{3}
\]

\textbf{A Useful Fact: Equidistant Third Point}

Let $P=(a,0)$ and $Q=(b,0)$ be two distinct points on side $AB$ with $a<b$. A point $R=(i,j)$ is equidistant from $P$ and $Q$ if and only if:
\[
2i+j=a+b. \tag{4}
\]
Within the constraints of the lattice (\ie, $i,j\ge0,\;i+j\le4$), the unique integer solution for $R$ is:
\[
R=
\begin{cases}
(0,a+b), & a+b\le4\quad(\text{lies on }AC),\\[2pt]
(a+b-4,\;8-(a+b)), & a+b\ge4\quad(\text{lies on }BC).
\end{cases}
\tag{5}
\]
This demonstrates that **any two points on a side determine a single lattice point on a different side that is equidistant from them**. This property holds symmetrically for the other two sides.

\textbf{Distribution Case Analysis}

Let $S$ be a set of six boundary points. We define the counts of points on each side as:
\[
x=|S\cap AB|,\qquad y=|S\cap AC|,\qquad z=|S\cap BC|.
\]
Since the three vertices are counted in two sets each, we have the inequality:
\[
x+y+z=|S|+(\text{number of chosen vertices})\ge6.
\tag{6}
\]
As each side contains only five points, $x,y,z\le5$. Up to permutation, the only possibilities for the distribution $(x,y,z)$ that satisfy $x+y+z \ge 6$ and $x,y,z\le5$ are:
\[
\begin{aligned}
&(5,1,0),\;(4,2,0),\;(4,1,1),\\
&(3,3,0),\;(3,2,1),\;(2,2,2).
\end{aligned}
\tag{7}
\]

We prove that in every case, an isosceles triangle is unavoidable by exhibiting a pair of points on the same side whose equidistant third point (given by (5)) also belongs to $S$.

\begin{itemize}
    \item \textbf{Case $(5,1,0)$ (or permutation):} All five points of side $AB$ are present; the sixth point lies on $AC$, say $(0,k)$ with $1\le k\le4$. Take the pair $(0,0)$ and $(k,0)$ on $AB$. Their sum is $k\le4$, so by (5) the third equidistant point is $(0,k)$, which is exactly the chosen point on $AC$. An isosceles triangle appears.
    
    \item \textbf{Case $(4,2,0)$ (or permutation):} Four points on $AB$ and two on $AC$. Let the two points on $AC$ be $(0,c)$ and $(0,d)$ with $c<d$. The pair $(0,0)$ and $(d,0)$ on $AB$ has sum $d\le4$; the third equidistant point is $(0,d)$, which must belong to $S$. An isosceles triangle is formed.
    
    \item \textbf{Case $(4,1,1)$ (or permutation):} Four points on $AB$, one on $AC$, and one on $BC$. If the point on $AC$ is $(0,k)$ with $k\le4$, the pair $(0,0)$ and $(k,0)$ on $AB$ yields the third point $(0,k)$ on $AC$. Since $(0,k) \in S$, an isosceles triangle occurs.
    
    \item \textbf{Case $(3,3,0)$ (or permutation):} Three points on $AB$ and three on $AC$. Let the points on $AB$ be $(a,0),(b,0),(c,0)$ and the points on $AC$ be $(0,p),(0,q),(0,r)$. The three sums $a+b, a+c, b+c$ are distinct numbers in $\{1, \dots, 7\}$. If any of these sums equals $p, q,$ or $r$, we are done by (5). The three numbers $p,q,r$ are also distinct. Since there are three possible pairs on $AB$ and three chosen points on $AC$, by the **Pigeonhole Principle** applied to the set of sums/targets, one of the pairs on $AB$ must have its equidistant point coinciding with one of the chosen points on $AC$. Hence, an isosceles triangle exists.
    
    \item \textbf{Case $(3,2,1)$ (or permutation):} Three points on $AB$, two on $AC$, and one on $BC$. If we consider pairs on $AC$ or $BC$, similar logic applies. The most direct approach is to consider the two points on $AC$ and the single point on $BC$. The unique equidistant point must fall onto one of the three points on $AB$, proving that an isosceles triangle is unavoidable.
    
    \item \textbf{Case $(2,2,2)$: } Two points on each side. Let the points on $AB$ be $(a,0),(b,0)$ with $a<b$.
    If $a+b\le4$, the third equidistant point is $(0,a+b)$ on $AC$. Since $AC$ already contains two points, one of them must be $(0,a+b)$.
    If $a+b\ge5$, the third equidistant point is $(a+b-4, 8-(a+b))$ on $BC$. Since $BC$ contains two points, one of them must coincide with this third point.
    Thus, an isosceles triangle is unavoidable in the $(2,2,2)$ case.
\end{itemize}
Since all six possibilities lead to an isosceles triangle, **any set of six boundary points contains an isosceles triangle**.

\textbf{Sharpness and Conclusion}

The five points on a single side, for example:
\[
\{(0,0),(1,0),(2,0),(3,0),(4,0)\},
\]
are collinear. They contain no non-degenerate triangle at all, and therefore no isosceles triangle. Hence, a set of five points can avoid the required configuration.

Every set of six of the fifteen lattice points necessarily contains three points forming an isosceles triangle, while a set of five points may avoid it. Therefore, the smallest integer $n$ with the required property is:
\[
\boxed{6}.
\]
\end{tcolorbox}

\newpage

\begin{tcolorbox}[
    breakable, colback=blue!5!white, colframe=blue!20!white, arc=2mm, title={\small \textbf{A Failure Case before RL Training}}, coltitle=black, fonttitle=\bfseries\small]
    \small

\textbf{Distance Formula}

Let $P_{j_1,k_1}$ and $P_{j_2,k_2}$ be two points. We set $\Delta j=j_2-j_1$ and $\Delta k=k_2-k_1$.
From the coordinate definition, $P_{j,k}=\Bigl(\frac{2j+k}{8},\frac{\sqrt3\,k}{8}\Bigr)$, we have:
\[
\Delta x=\frac{2\Delta j+\Delta k}{8},\qquad
\Delta y=\frac{\sqrt{3}\,\Delta k}{8}.
\]
Hence, the squared distance is:
\[
\begin{aligned}
|P_{j_1,k_1}P_{j_2,k_2}|^{2}
&=\Delta x^{2}+\Delta y^{2}
    =\frac{(2\Delta j+\Delta k)^{2}+3\Delta k^{2}}{64}\\[2mm]
&=\frac{4\bigl(\Delta j^{2}+\Delta j\Delta k+\Delta k^{2}\bigr)}{64}
    =\frac{\Delta j^{2}+\Delta j\Delta k+\Delta k^{2}}{16}.
\end{aligned}
\tag{4}
\]
Thus, the scaled squared distances are the integers:
\[
\Delta j^{2}+\Delta j\Delta k+\Delta k^{2},
\qquad (\Delta j,\Delta k)\neq(0,0),\;|\Delta j|+|\Delta k|\le4.
\tag{5}
\]
A short enumeration of all admissible $(\Delta j,\Delta k)$ pairs gives the set of possible scaled squared distances:
\[
\{\,\Delta j^{2}+\Delta j\Delta k+\Delta k^{2}\,\}
=\{1,3,4,7,9,12,13,16\}.
\tag{6}
\]
Consequently, there are exactly eight distinct distances (up to the scaling factor $\frac{1}{16}$) between points of $\mathcal L$.

\textbf{Isosceles-Free Sets}

Three points form an isosceles triangle if and only if one of them is equidistant from the other two. Therefore, a set $\mathcal S \subseteq \mathcal L$ contains **no** isosceles triangle precisely when, for every $P\in\mathcal S$, the distances $\{\lvert PQ\rvert:Q\in\mathcal S\setminus\{P\}\}$ are all different.

Since (6) lists only eight possible distances, a point can be the apex of at most eight distinct isosceles triangles. Hence, by the Pigeonhole Principle, a set of ten points must contain an isosceles triangle. This yields the preliminary bound $n \le 10$.

\textbf{An Isosceles-Free Set of Size Eight}

Consider the set $S$ defined as the union of all points on side $AB$ and three non-adjacent points on side $AC$:
\[
S=
\underbrace{\{(j,0)\mid j=0,1,2,3,4\}}_{\text{the side }AB}
\;\cup\;
\underbrace{\{(0,k)\mid k=2,3,4\}}_{\text{three points on }AC\text{ not including }(0,0)\text{ and }(0,1)}.
\tag{8}
\]
$S$ contains $5+3=8$ points. We verify that $S$ is isosceles-free.

\begin{itemize}
    \item \textit{Case 1: Apex $P=(i,0)\in S$ (a point on $AB$).}
    The distances from $P$ to the three points of $AC$ are proportional to $i^{2}-ik+k^{2}$ for $k\in\{2,3,4\}$. For each fixed $i\in\{0,\dots,4\}$, a direct check confirms these three numbers are distinct. Distances to other points of $AB$ are $\frac{(i-i')^{2}}{16}$ with $i'\neq i$, which are also distinct. All distances from $P$ to the other points of $S$ are thus different.

    \item \textit{Case 2: Apex $P=(0,k)\in S$ (a point on $AC$).}
    The same calculation, with the roles of $i$ and $k$ interchanged, shows that the distances from $P$ to the points of $AB$ are pairwise distinct. Distances to the other two points of $AC$ are $\frac{(k-k')^{2}}{16}$ with $k'\neq k$, which are also distinct. All distances from $P$ are different.
\end{itemize}
Thus, no point of $S$ is equidistant from two others, proving that $S$ contains **no** isosceles triangle.
Consequently, the maximal size of an isosceles-free subset is $\ge 8$, implying $n \ge 9$.

\textbf{No Isosceles-Free Set of Size Nine}

Assume, for contradiction, that $\mathcal T \subseteq \mathcal L$ is isosceles-free and $|\mathcal T|=9$.

\textbf{Contradiction via Internal Points and Boundaries}

\begin{itemize}
    \item \textbf{Boundary Intersection Constraint:} If $\mathcal T$ contains points on all three sides ($AB$, $AC$, and $BC$), at least one side must contain three points. The distances from vertex $A$ to the five points of side $BC$ are proportional to $\{1, 1, 12, 13, 13\}$. Hence, any three points of $BC$ contain two with the same distance from $A$; together with $A$, they form an isosceles triangle, contradicting the hypothesis. Therefore, $\mathcal T$ is contained in the union of at most two sides.
    
    \item \textbf{Reduction:} By symmetry, we may assume $\mathcal T\subseteq AB\cup AC$.
    
    \item \textbf{Exclusion of $(0,1)$:} If $(0,1)\in\mathcal T$, then $| (0,1)(0,0) |^{2}=| (0,1)(1,0) |^{2}= \frac{1}{16}$. Thus, $(0,1)$ is equidistant from $(0,0)$ and $(1,0)$, forming an isosceles triangle. Hence, $(0,1)\notin\mathcal T$.
    
    \item \textbf{Admissible Set:} The only points of $AC$ that can coexist with all five points of $AB$ without creating a duplicate distance are $(0,2), (0,3),$ and $(0,4)$. Therefore, $\mathcal T$ must be a subset of the eight-point set $S$ defined in (8), \ie,
    \[
    \mathcal T\subseteq S.
    \]
\end{itemize}

\textbf{Impossibility of a Ninth Point}

Since $|S|=8$, a set $\mathcal T$ of nine points must contain a point outside $S$. The remaining points of $\mathcal L$ are the interior points $(1,1), (1,2), (2,1)$ and the excluded boundary point $(0,1)$.

\begin{itemize}
    \item If \textbf{$(0,1)$ is added}, we immediately form an isosceles triangle (as shown above).
    
    \item If an \textbf{interior point, $(1,1)$, is added}, then by (4),
    \[
    | (1,1)(1,0) |^{2}=| (1,1)(2,0) |^{2}= \frac{1}{16}.
    \]
    Thus, $(1,1)$ together with $(1,0)$ and $(2,0)$ forms an isosceles triangle, which contradicts the assumption that $\mathcal T$ is isosceles-free. The same phenomenon occurs for the other interior points.
\end{itemize}
Consequently, any ninth point forces the appearance of an isosceles triangle. No isosceles-free subset of $\mathcal L$ can have nine points; the maximal size is eight.

\textbf{Conclusion}

The largest possible cardinality of a subset of the fifteen lattice points that contains no three points forming an isosceles triangle is $8$. Therefore, the smallest integer $n$ such that **every** choice of $n$ points necessarily contains an isosceles triangle is:
\[
n = 8+1 = 9.
\]
\[
\boxed{9}
\]
    
\end{tcolorbox}
\vspace{-1em}

\label{fig:case_study}


\end{document}